\documentclass{article}

\usepackage[preprint]{neurips_2026}

\usepackage[utf8]{inputenc}
\usepackage[T1]{fontenc}
\usepackage[hidelinks]{hyperref}   % hidelinks = no colored boxes around links
\usepackage{url}
\usepackage{booktabs}
\usepackage{amsfonts}
\usepackage{amsmath}
\usepackage{amssymb}
\usepackage{nicefrac}
\usepackage{microtype}
\usepackage{xcolor}
\usepackage{graphicx}
\usepackage{subcaption}

\title{Cross-Modal Forgetting and Transfer in Unified Multimodal Post-Training} 

\workshoptitle{NeurIPS 2026 Workshop on Continual Learning in the Era of
Foundation Models and Embodied Agents}

\author{%
  Francesco Colasurdo\thanks{Corresponding author.} \\
  Eindhoven University of Technology \\
  \And
  Ivan Slobozhan \\ Eindhoven University of Technology \\
  \And
  Joaquin Vanschoren \\ Eindhoven University of Technology \\
  \And
  Anna Vettoruzzo \\ Eindhoven University of Technology \\
}

\begin{document}
\maketitle

\begin{abstract}
Unified multimodal models answer questions about images and also generate images with one
shared set of weights. This creates a post-training risk that single-purpose models do not
have: fine-tuning one capability can damage another that works in a different modality.
We measure this cross-modal forgetting for LoRA post-training of Janus-Pro-7B. Every
target is chosen by how poorly the base model already handles it, from a poetic register
to a language it barely speaks, so that retention is tested against a real intervention;
we then evaluate the capabilities we never trained, first after a single run and then
after chaining three. A single run looks entirely safe: nothing we did not train degrades
on four public benchmarks or in free-form English, even when the adapter cuts held-out Welsh perplexity by $76\%$ with a relative weight change under $2\%$.
 That safety belongs to
the run, not to the recipe. Chaining three runs destroys the capability learned one run
earlier and raises general-text perplexity by one to two orders of magnitude above the never-adapted model, on the very measures that stayed flat before. The cause is in the
objective, which constrains only the modality being trained and leaves the other head unconstrained:
cutting the learning rate to one fifth does not prevent the collapse, while anchoring the other head removes the damage to the capabilities we never trained, at no cost to what is learned. What is learned
also stays inside its own modality, with Welsh acquired from text closing only $3\%$ of
the English--Welsh gap in image generation.
\end{abstract}

\section{Introduction}
Continual learning asks how a model can learn something new without losing what it
already knows \citep{mccloskey1989catastrophic, french1999catastrophic}. For foundation
models this is now a practical question. Post-training is the usual way to specialize a
released checkpoint, and every such update can erode the pretrained abilities that made
the checkpoint worth using \citep{kirkpatrick2017ewc, li2017lwf}.

Unified multimodal models (here always vision-language models) give this question a new form. They do two jobs with one set of
weights: image-text understanding (image and text in, text out) and generation (text in, image out).
When post-training specializes one of them, damage can appear not only in related tasks
in the same modality but also in tasks in the \emph{other} modality that share the
weights. We call this phenomenon \emph{cross-modal forgetting}. It cannot happen in a single-purpose model
such as a text-to-image diffusion model, because its weights serve a single output pathway; in a unified model they serve two, so it is a direct test of the ``one backbone, many capabilities'' idea.

Such models are usually specialized with LoRA \citep{hu2022lora}, the standard post-training tool, which is often assumed to leave the untrained capabilities intact because it changes few parameters. What is known about that assumption was measured within one modality \citep{biderman2024lora, luo2023empirical, zhai2023investigating}, and the work on understanding-generation conflict in unified models
changes the architecture to avoid the problem rather than measuring it. How much damage a
single unmodified LoRA run does, which untrained capabilities it reaches, and where the
line lies between a run that stays contained and one that does not, are the questions we
measure.

Measuring this credibly requires a careful choice of target. If the base model already
handles the fine-tuning data well, the adapter barely needs to change the weights.
Finding ``no forgetting'' in that case means nothing, because a change that small could
not have damaged anything. We therefore pick every target by how poorly the base model
handles it, using one scale that applies to all targets. We also report how much each
adapter changed the weights, so a reader can check that the intervention was real before
trusting that nothing broke.

\paragraph{Contributions.}
\textbf{(i)} We make measurable the condition under which ``PEFT does not forget'' says anything at all: the target must be outside what the base model can already do.
Every target is scored by the base model's perplexity on it against a fixed control, every
adapter is reported with the size of its weight change, and the instruments are
calibrated.
\textbf{(ii)} Under that condition a \emph{single} fine-tuning run changes only what it
trained. Across five targets, LoRA ranks $2$ to $64$, and KL weights
$\lambda\in[0,1.5]$, the untrained capabilities stay at the base level on the four
benchmarks and in the image-generation distribution; no adapter degrades free-form
English, and this holds even for the Welsh language, which the base model barely knew ($40.7\to9.6$). The first cost appears in grounded visual understanding, at the strongest settings. We call this containment: the damage stays inside the capability being trained. \textbf{(iii)} That containment belongs to a single run, not to the recipe: the regularizer is computed on the modality being trained and leaves the other head unconstrained. Chaining three runs, each starting from the previous run's adapter, destroys the capability learned one run earlier (held-out perplexity for Welsh $9.56\to4634$, for medicine $8.67\to13513$) and raises general-text perplexity above the never-adapted model by $17\times$ in one order and $142\times$ in the other, on the measures that stay flat after one run. At one fifth of the learning rate the damage is seven times smaller but still there:
the sequence causes the collapse; the learning rate sets how large it is.
Anchoring the other head as well removes the damage to the capabilities we never trained, at no cost to acquisition, though it does not restore the task learned one run earlier.
\textbf{(iv)}  The learned language does not transfer across modalities. 
It improves image generation from Welsh prompts only marginally, $+0.48\pm0.24$ CLIPScore,
$3\%$ of the English--Welsh gap, and leaves Welsh visual question answering at chance.

\section{Related Work}
Continual learning is the conflict between keeping what a model knows and learning
something new with the same weights \citep{mermillod2013stability, delange2022survey, wang2024clsurvey}. 
Three kinds of remedy recur: \emph{regularization} toward a reference, in weight space (L2-SP
\citep{xuhong2018l2sp}, elastic weight consolidation, EWC \citep{kirkpatrick2017ewc}) or in output space (distillation \citep{hinton2015distilling}, Learning-without-Forgetting, LwF \citep{li2017lwf});
\emph{replay and constraint} \citep{lopezpaz2017gem, chaudhry2019agem}; and
\emph{parameter isolation} \citep{rusu2016progressive, mallya2018packnet,
serra2018hat}. Our setup borrows from the first and the third: a KL penalty keeps the
adapter's outputs close to the frozen base model, and LoRA itself is a mild form of
parameter isolation. A frozen reference cannot protect a task the base model never had,
which is why LwF anchors to the previous model rather than the initial one and why EWC
accumulates its Fisher across tasks \citep{li2017lwf, kirkpatrick2017ewc}.

Forgetting in large pretrained models is real but depends on scale and method
\citep{ramasesh2022scale}; parameter-efficient methods learn less and forget less than
full fine-tuning \citep{biderman2024lora}, and little is gained from rank beyond a small
value \citep{zhang2024scalingmeetsfinetuning}. It is documented for language models
under continual fine-tuning \citep{luo2023empirical} and for multimodal ones
\citep{zhai2023investigating}, but always within one modality or within tasks of the
same kind. Methods built for sequential LoRA keep each task in its own subspace, either
orthogonal to the previous ones \citep{wang2023olora} or chosen so that the new task interferes least with them \citep{liang2024inflora}; others route between several adapters
\citep{dou2024loramoe, wu2024mole, li2024mixlora}. All of them take the interference as
given.

Unified models make the question pressing. Fully tokenized models \citep{chameleon2024,
bachmann20244m21, lu2024unifiedio2} put every modality in one token sequence; Janus and
Janus-Pro \citep{wu2024janus, januspro} keep separate visual encoders over one shared
backbone. LoRA has been applied to such models before \citep{chern2024anole,
serra2024narrowgate, wang2025vora}, but as a tool, not as the object of study. We make it the object of study, asking what one unmodified LoRA run costs the capabilities the checkpoint already had. The closest
work is MoDE \citep{wei2025mode}: it fine-tunes Chameleon and Janus-Pro on a sequence of
instruction-tuning tasks with LoRA, finds forgetting within and across modalities, and
proposes per-modality routing that prevents it. MoDE protects the tasks the model 
was trained on in sequence; we protect pretrained abilities
that were never a training task at all, which a practitioner inherits with the checkpoint
and no task benchmark scores. That difference sets our protocol (\S\ref{sec:setup}). Where
the two overlap, our sequential results agree with theirs. We change neither the architecture nor
the adapter.

\section{Setup}
\label{sec:setup}
\paragraph{Model, capabilities, and adapter.}
We use Janus-Pro-7B \citep{januspro}: one shared transformer backbone with separate
encoders for understanding (SigLIP) and generation (a VQ tokenizer), giving three
capabilities: text-to-text language modeling, image-text understanding and text-to-image
generation (Appendix~\ref{app:impl}). We insert LoRA into the backbone
($q,k,v,o,\texttt{gate},\texttt{up},\texttt{down}$) with $\alpha=2r$ fixed across the rank
sweep. Targets are selected by module-name suffix, so the same pattern also catches the $q,k,v$ projections of the SigLIP encoder: of the $282$ adapted modules, $210$ are in the backbone and $72$ in SigLIP. The VQ tokenizer, the image head and everything else stay frozen. The SigLIP adapters turn out never to be updated: neither direction encodes an image for understanding, so their $B$ matrices are exactly zero in every adapter we trained, and the effective adaptation is the $210$ backbone modules. Both directions therefore train the shared backbone alone. Understanding reads from that same backbone, so it is exposed to every update we make even though its own encoder is untouched, and it still did not degrade (\S\ref{sec:gen}, \S\ref{sec:text}).

\paragraph{Keeping the adapter close to the base model: a KL penalty.}
In both directions we can optionally hold the adapter near the frozen base model. With
$p_{\text{adapter}}$ and $p_{\text{base}}$ the next-token distributions of the same model
with the adapter on and off, and $m$ the head being trained, the loss is
\begin{equation}
\mathcal{L}=\mathrm{CE}\big(p^{m}_{\text{adapter}},y\big)
+\sum_{m'}\frac{\lambda_{m'}}{T_{m'}}\sum_{t=1}^{T_{m'}}\mathrm{KL}\!\big(p^{m'}_{\text{adapter}}(\cdot\mid x_{<t})\,\big\|\,\mathrm{sg}[p^{m'}_{\text{base}}(\cdot\mid x_{<t})]\big),
\label{eq:kl}
\end{equation}
where $m'$ ranges over both heads: text logits, and generation-head logits at image
positions. The reference costs nothing extra: it is the same model with the adapter
disabled, held by a stop-gradient, so no separate teacher is stored. $\lambda_{m}=0$ is
plain fine-tuning and larger $\lambda_{m}$ keeps the adapter closer to the base. We match
output distributions rather than weights because the output distribution is what we want
to preserve. Compared to L2-SP, logit-MSE and the reconstruction objective of Anole
\citep{chern2024anole} at matched update size, the KL penalty learns the target best, and
none of the five disturbs the untrained capabilities (Appendix~\ref{app:reg}).
Every run below sets $\lambda_{m'}=0$ for $m'\neq m$: both terms are then computed on the
head being trained, and the other head receives no gradient at all. The single exception
is the intervention reported below, which sets the text term to
$\lambda_{\text{txt}}=1$ during an image run.

\paragraph{Evaluation after adaptation.}
We evaluate the capabilities we did not train with four standard benchmarks that are
scored by log-likelihood alone, so they are deterministic and need no judge: WikiText-2
perplexity \citep{wikitext}, MMLU \citep{mmlu}, POPE \citep{pope} and MME \citep{mme}.
Our base POPE accuracy ($0.877$) matches the published Janus-Pro-7B value ($0.874$), which
validates the evaluation harness. POPE is therefore the externally validated benchmark.
For MME we report the full total over all $14$ subtasks ($n{=}2374$) and our base scores
$1785$; the Janus-Pro paper \citep{januspro} reports $1567.1$ on the perception subset
alone, so the two totals are not comparable and we compare MME only between our own runs.
Two evaluation configurations are used throughout, a full one and a subsampled one
(Table~\ref{tab:evalsizes} in Appendix~\ref{app:impl}).

Because likelihood benchmarks never make the model generate freely, we add two metrics of
generated output: MAUVE \citep{pillutla2021mauve} for text, and, for image generation
under a text adapter, the CMMD \citep{cmmd} between the adapter's images and the base
model's images on a fixed set of prompts, which we call Gen-CMMD.
A distance near zero means nothing without a scale, so we calibrate Gen-CMMD on the image
adapter, whose images are visibly restyled: $18.7\times10^{-3}$ against $1.8\times10^{-3}$
for the text adapters. KID is positive for the former and slightly negative for the latter.

\paragraph{What we count as retained, and how much can be lost.}
Accuracy on a multiple-choice benchmark mixes real knowledge with lucky guesses, and only
the knowledge can be lost: with $k$ options the losable part is
$a_{\text{true}}=(ka-1)/(k-1)$, a chance correction we take from
\citet{harmon2025postforgetting}, which for the base model gives $32.3\%$ on MMLU
($k{=}4$), $75.5\%$ on POPE and $57.6\%$ on MME ($k{=}2$). Applying it to every adapter
leaves our conclusions unchanged: no adapter loses a detectable share of the losable knowledge on any benchmark (\S\ref{sec:gen}, \S\ref{sec:text}; Appendix~\ref{app:headroom}). We call a capability \emph{retained} when the adapter's score is within evaluation noise of the base model's. A large movement is a behavioral change whatever its sign.
General-text perplexity moves under the text adapters and the medicine adapter shifts
MAUVE-vs-base; we treat both as changes rather than evidence of stability (\S\ref{sec:text}).

\paragraph{Choosing targets the base model handles poorly.}
We score each candidate target by $\times$control: the base model's perplexity on that
corpus divided by its perplexity on a fixed English control corpus. A higher value means
the corpus is further from what the base model already handles well. On this scale the three text targets are poetry (Project Gutenberg) at $4.0\times$, Welsh
(\texttt{wikimedia/wikipedia}, Welsh subset) at $2.3\times$, and medical abstracts
(\texttt{ccdv/pubmed-summarization}) at $1.2\times$: a register, a language and a domain. Further design choices are in Appendix~\ref{app:langprobe}. For images we use a narrow anime style
(\texttt{lambdalabs/naruto-blip-captions}, $1{,}141$ pairs), which the base model renders
poorly, and black-and-white line sketches (\texttt{zoheb/sketch-scene}). An image style
is scored against real images in that style, with CMMD \citep{cmmd} as the main metric,
supported by KID \citep{binkowski2018demystifying}, DINOv2 \citep{dinov2}, CLIP-IQA
\citep{clipiqa} and CLIP directional similarity \citep{stylegannada}. All numbers come from a single training seed, except the image run of each sequence in \S\ref{sec:chain}, which was repeated at a second initialization.
Full hyperparameters, and the standard errors, which measure
evaluation-set sampling noise at fixed weights, are in Appendix~\ref{app:impl}. We fine-tune one capability at a time and evaluate the others, reporting for each target how
well it is learned, what happens to the untrained capabilities, and any shift in the other modality.

All training and evaluation code, run configurations and the exact evaluation subsamples are available at \url{https://github.com/kaj04/multimodal-peft}.

\section{Results}

\subsection{Fine-tuning image generation}
\label{sec:gen}
We fine-tune image generation with the configuration that learned the style best in an incremental ablation ($r=48$, LoRA on attention and MLP, $\lambda=0.5$); that ablation, and the images the configuration produces, are in Appendix~\ref{app:ablation} (Figure~\ref{fig:janus_qualitative_comparison}).
Table~\ref{tab:rank} varies the rank from $2$ to $64$: learning of the style improves and
then levels off (CMMD $0.139\to0.097$, directional similarity $0.143\to0.158$, with most of the gain arriving by rank $16$ to $32$), while every untrained capability stays at the base level, with WikiText
within $0.07$ of the base and MMLU, POPE and MME within sampling noise. We repeat this with
line sketches (Appendix~\ref{app:sketch}), which leave the largest footprint of all five
targets and are one of the three settings described in \S\ref{sec:boundary}.

\subsection{Fine-tuning language modeling}
\label{sec:text}
\paragraph{Varying the KL weight.}
At rank $16$ on poetry, increasing $\lambda$ trades learning for preservation
(Table~\ref{tab:textkl}), and MME shows a small dip at $\lambda{=}0$ that the penalty
removes. Because the target here is text, the cross-modal question is what happens to images: Gen-CMMD stays near $1.7\times10^{-3}$ at every $\lambda$, including $\lambda{=}0$, where the text objective is completely
unconstrained. Specializing the text stream does not measurably move the image-generation
distribution even with no KL penalty at all.

\paragraph{A domain and a language are learned without collateral damage.}
Poetry and medicine are both within the base model's competence (poetic English, medical
English), so the absence of forgetting there could still be explained by a weak
intervention. We therefore add a target the base model barely speaks: Welsh.
Table~\ref{tab:targets} reports both. Medicine is learned strongly. Held-out PubMed
perplexity goes from $12.8\to8.4$, and on the $\times$control scale the domain moves from
$1.2\times$, harder than generic English, to $0.79\times$, easier. Welsh is learned more
strongly still. Held-out perplexity is reduced from $40.7$ to $9.6$, a $-76\%$ change, and given a
Welsh opening the adapter continues in Welsh where the base model slips into English or into its
assistant persona; continuations for all three targets are in Appendix~\ref{app:textqual}. In both cases, the untrained capabilities do not degrade on any of the four benchmarks (Table~\ref{tab:targets}), and Gen-CMMD stays at the same level as for poetry ($1.77$ for medicine and $1.83\times10^{-3}$ for Welsh).

How far a target sits from what the base model already handles predicts neither outcome.
On the common $\times$control scale Welsh ($2.3\times$) sits between poetry ($4.0\times$) and medicine
($1.2\times$), yet it gives the largest perplexity reduction and the largest weight change
($1.9\%$, versus $1.3\%$ and $1.7\%$). The size of the change follows the perplexity
reduction ($-25\%$, $-34\%$, $-76\%$) and not the $\times$control ranking, under which poetry is the most distant target and moves the weights least. What was learned is also
weaker than a language: the samples in Appendix~\ref{app:textqual} remain locally
repetitive under greedy decoding, and the held-out text comes from the same Wikipedia
subset as the training text, so the drop mixes the language with that source's encyclopedic
register. We therefore claim the adapter learned a corpus, and judge retention against that
weaker achievement.

\paragraph{Free-form generation.}
MAUVE \citep{pillutla2021mauve} closes the gap the likelihood benchmarks leave open
(Table~\ref{tab:mauve}, Appendix~\ref{app:mauve}). The image adapters write free English essentially identical to the base model (MAUVE-vs-base $0.95$/$0.92$). Within text, no adapter's free English is farther from human text than
the base model, and the text adapters are closer ($0.60$--$0.70$). The medicine adapter is the exception: its MAUVE-vs-base of $0.21$ is a large distributional change, which by our definition is a behavioral
change and not retention. On the target side the text adapters move much closer
to their target distributions than the base model (poetry $0.45\to0.80$, medical abstracts
$0.31\to0.91$, Welsh $0.14\to0.53$), confirming learning at the level of generated text and
not only in perplexity.

\paragraph{The first cost appears in grounded visual understanding.}
\label{sec:boundary}
Three settings at the strong end of fine-tuning move grounded visual understanding in the
same direction while knowledge and language stay flat: the poetry adapter at $\lambda=0$
(Table~\ref{tab:textkl}), poetry at rank $128$ (Appendix~\ref{app:textsweep}), and line
sketches, the hardest visual style (Appendix~\ref{app:sketch}). Each movement is on the order of one to two standard errors of a single seed, so we report the pattern as an indication of where
to look, not as a finding.

\subsection{How much do the adapters change the weights?}
\label{sec:dw}
A skeptic could say that nothing is forgotten because nothing is changed. The size of the
weight changes rules that out. We summarize each change $\Delta W=(\alpha/r)BA$ by
$\lVert\Delta W\rVert_F/\lVert W\rVert_F$, a task-vector magnitude \citep{ilharco2023task}:
small but real, $0.6$ to $1.9\%$, ordered by both knobs in both directions
(Table~\ref{tab:dw-rank}). Changes of under $2\%$ already shift behavior substantially:
poetry perplexity goes from $32.7$ to $23.8$ and visible style transfer appears in
Figure~\ref{fig:janus_qualitative_comparison}.

\subsection{What is learned does not cross modalities}
\label{sec:transfer}
After learning Welsh through text, can the model \emph{use} Welsh in the visual
capabilities? This is a question about transfer, not retention: the base model is expected
to fail on Welsh because it barely knows the language, so any gain must come from the
adapter (protocol in Appendix~\ref{app:xmodal}). The architecture leaves the door open,
since Janus generates images through the same weights the adapter modifies. The training
does not: the adapter saw only text tokens and the objective was plain language modeling on
Welsh Wikipedia, not instruction following in Welsh.

The answer is essentially no (Table~\ref{tab:xmodal}). When images are generated from
Welsh prompts, the adapter improves CLIPScore against the same English reference only
marginally over the base model ($10.7\to11.1$, $+0.48\pm0.24$, $18/30$ prompts improved),
closing just $3\%$ of the English--Welsh gap ($15.7$). A fixed-seed control
(Appendix~\ref{app:xmodal}) shows that Welsh text conditions the generator but conveys no
content: unrelated Welsh prompts give different images, all of them close to what an
empty prompt produces. On Welsh visual question answering we detect no transfer: on $28$
balanced yes/no items the base model scores $53.6\%$ and the adapter $46.4\%$, both
indistinguishable from chance ($95\%$ Wilson intervals $[36,70]$ and $[30,64]$) against
$100\%$ in English for both.

The probe measures less than it appears to. A text-only control settles what: replacing
the image with a single English sentence and keeping the $28$ questions verbatim, the
base model still scores $53.6\%$ on the Welsh questions against $100\%$ on the English
ones, and the adapter $57.1\%$. The base model's English--Welsh gap is the same with no image present
as with one, so probe~(B) measures Welsh instruction following and not the visual
pathway; the item-level pattern behind it is in Appendix~\ref{app:xmodal}. With $n{=}28$
the Wilson intervals are in any case wide enough that a fifteen-point gain would go
undetected, so we read probe~(B) as an upper limit rather than a null.

The locality of the update explains both halves of the single-run result: what stays
protected, and what fails to transfer. The KL penalty shapes it but does not cause it.
Five objectives that differ by a factor of two in how well they learn the style all leave
every untrained capability within evaluation noise (Appendix~\ref{app:reg}), and Gen-CMMD
is flat at every $\lambda$, including zero. Locality at this scale is a property of
low-rank adaptation, and the penalty earns its place by improving acquisition. Across runs it cannot help at all, because it is computed on the modality being trained. That is what \S\ref{sec:chain} measures.

\section{Containment holds for one run but not for a sequence}
\label{sec:chain}
A checkpoint is rarely post-trained only once. We therefore repeat the same recipe in
sequence, changing one thing: each run starts from the previous run's adapter instead of
from the base model. Concretely, the same LoRA matrices keep training. The adapter is
injected once, the previous run's weights are loaded into it, and the optimizer, the
scheduler and the epoch counter are reset; no new parameters are added and nothing is
merged into the base.

We run two chains of three runs in opposite orders. In chain~A, run~1 is Welsh, run~2 the anime image
style and run~3 medicine; in chain~B, run~1 is medicine, run~2 the same image style and run~3 Welsh.
The image run is held constant in the middle, and each text target comes first in one chain and last in
the other. Every run uses one configuration throughout ($r{=}32$, $\lambda{=}0.5$, three
epochs for text and six for image), which for medicine means $\lambda{=}0.5$ rather than
the $\lambda{=}0.25$ of Table~\ref{tab:targets}, and for the image run $r{=}32$ rather
than the $r{=}48$ of \S\ref{sec:gen}: trained from the base it reaches $8.67$
here against $8.4$ there. After every run we repeat the full evaluation.
Table~\ref{tab:chain} reports the result.

We propose no continual-learning method and compare none; reusing one unchanged configuration
is what makes the chain comparable to the rest of the paper. The one
intervention we report below tests an explanation rather than offering a remedy.

\begin{table}[t]
\caption{\textbf{Chained fine-tuning: the retention matrix.} Two orders, A and B; bold marks the target trained in that run. All rows use the subsampled configuration ($570$/$1000$/$600$), not the full one of Tables~\ref{tab:rank} and~\ref{tab:targets}. Style CMMD is against real style images, where a standalone image adapter reaches $0.108$. $^{\dagger}$standalone medicine run. Single seed except run~2 of each chain.}
\label{tab:chain}
\centering\footnotesize
\setlength{\tabcolsep}{3.5pt}
\begin{tabular}{lrrcrrrrr}
\toprule
& \multicolumn{2}{c}{Target PPL $\downarrow$} & Style & \multicolumn{4}{c}{Untrained capabilities} & $\lVert\Delta W\rVert$ \\
\cmidrule(lr){2-3}\cmidrule(lr){5-8}
After step & Welsh & Medicine & CMMD $\downarrow$ & WikiText & MMLU & POPE & MME & /$\lVert W\rVert$ \\
\midrule
\emph{Base (subsampled)} & $40.7$ & $12.8^{\dagger}$ & $\approx.24$ & $13.71$ & $.481$ & $.875$ & $1130$ & --- \\
\midrule
A1. Welsh (text)        & $\mathbf{9.56}$ & $12.07$ & $.237$ & $12.16$ & $.477$ & $.870$ & $1131$ & $1.90\%$ \\
A2. $+$ anime (image)   & $4634$ & $140.2$ & $\mathbf{.132}$ & $229.8$ & $.472$ & $.873$ & $1119$ & $2.76\%$ \\
A3. $+$ medicine (text) & $40.80$ & $\mathbf{8.67}$ & $.151$ & $12.14$ & $.477$ & $.873$ & $1101$ & $3.33\%$ \\
\midrule
B1. medicine (text)     & $40.77$ & $\mathbf{8.67}$ & $.242$ & $11.87$ & $.481$ & $.868$ & $1123$ & $1.48\%$ \\
B2. $+$ anime (image)   & $279441$ & $13513$ & $\mathbf{.105}$ & $1950$ & $.440$ & $.819$ & $968$ & $2.39\%$ \\
B3. $+$ Welsh (text)    & $\mathbf{9.64}$ & $10.91$ & $.154$ & $12.37$ & $.465$ & $.873$ & $1136$ & $3.17\%$ \\
\bottomrule
\end{tabular}
\end{table}

\begin{figure}[t]
\centering
\setlength{\tabcolsep}{2pt}
\begin{tabular}{@{}cccc@{}}
\includegraphics[width=0.23\linewidth]{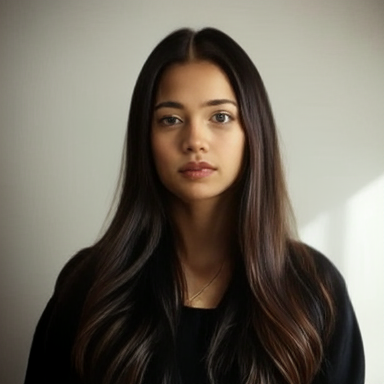} &
\includegraphics[width=0.23\linewidth]{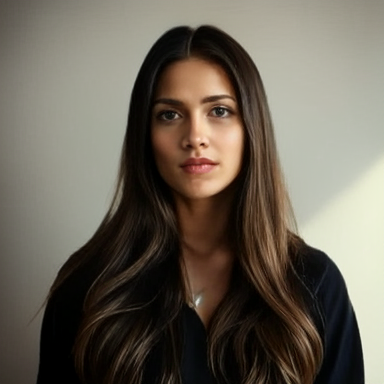} &
\includegraphics[width=0.23\linewidth]{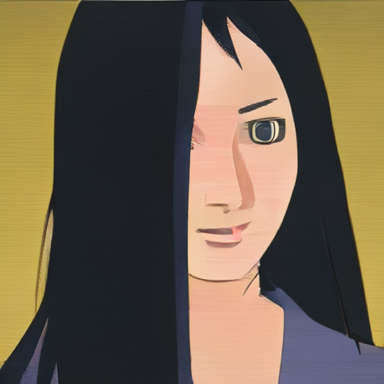} &
\includegraphics[width=0.23\linewidth]{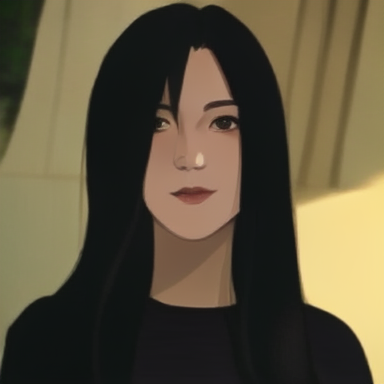} \\
\footnotesize Base & \footnotesize 1. Welsh & \footnotesize 2. $+$ anime &
\footnotesize 3. $+$ medicine \\
\end{tabular}

\vspace{3pt}
\footnotesize
\setlength{\tabcolsep}{5pt}
\begin{tabular}{@{}lp{0.76\linewidth}@{}}
\toprule
After step & \textit{``The most important scientific discovery of the twentieth century was''} \\
\midrule
Base            & \ldots the discovery of the structure of DNA. Hello, I am an AI language model\ldots \\
1. Welsh        & \ldots the discovery of the structure of DNA. Hello, I am an AI language model\ldots \\
2. $+$ anime    & \ldots the \texttt{\textbackslash n\textbackslash n\textbackslash n\textbackslash n}\ldots \emph{(whitespace to the token limit)} \\
3. $+$ medicine & \ldots the discovery of the structure of DNA ( deoxyribonucleic acid ) by Ja\ldots \\
\bottomrule
\end{tabular}
\caption{Chain~A at each step, both modalities, at a fixed seed. \textbf{Top:} images from ``a woman with long dark hair''; the text run leaves them untouched, the anime style arrives at run~2 and survives run~3. \textbf{Bottom:} greedy continuations of the prompt shown; after run~2 the language head emits only whitespace, and run~3 restores it. The same adapter that cannot produce a word of English still generates images in the learned style.}
\label{fig:chain-both}
\end{figure}

\paragraph{An image run is harmless from the base model and destructive after a text run.}
Fine-tuning image generation \emph{from the base model} leaves text untouched: WikiText
perplexity moves from $13.71$ to $13.78$ (Table~\ref{tab:rank}). The same fine-tuning
applied \emph{on top of} a text adapter destroys it. In chain~A held-out Welsh perplexity goes from
$9.56$ to $4634$; in chain~B held-out medicine perplexity goes from $8.67$ to $13513$; general English
follows, $12.16\to229.8$ and $11.87\to1950$. These are not returns to the starting
point: the base model scores $40.7$ on Welsh and $12.8$ on medicine, so the chained model
is two to three orders of magnitude \emph{worse} than one that was never fine-tuned.
Sequential-LoRA methods report forgetting as a return toward the pre-task level
\citep{wang2023olora, liang2024inflora}; the damaged task here ends far below it. The
perplexities also exceed the vocabulary size, which is not an evaluation artifact: after the image run,
greedy decoding emits only whitespace (Figure~\ref{fig:chain-both}), with $0.84$ of the
next-token mass on the newline where the base model puts $0.36$ on the correct word, and
the end-of-sequence token at rank $812$.

\paragraph{The damage is visible on the benchmarks.}
In chain~B the image run moves POPE
from $.868$ to $.819$ (about five standard errors), MME from $1123$ to $968$ (about four)
and MMLU from $.481$ to $.440$ (about two), and general-text perplexity rises by one to
two orders of magnitude in both chains. All three benchmarks therefore move when the damage is real, and their flatness everywhere else in this paper is a result, not a blind spot. The same
image run nonetheless leaves them flat in chain~A ($.477\!\to\!.472$, $.870\!\to\!.873$,
$1131\!\to\!1119$): what the run damages depends on what it was applied to, and the
identical training raises general-text perplexity six to nine times more in chain~B than in
chain~A ($11.87\to1950$ against $12.16\to230$; $6.2$ and $8.5$ in two runs).

\paragraph{The model still learns new tasks; it does not keep old ones.}
The chains do not leave a model that can no longer learn. The image style is acquired
normally at run~2 in both chains (CMMD $0.132$ and $0.105$ against $0.108$ for a
standalone image adapter), and the final text run reaches the quality of a single run from the base model: medicine ends chain~A at $8.67$, exactly its value from the base model, and
Welsh ends chain~B at $9.64$ against $9.56$. That last run also repairs what the image
run broke, returning general text to $12.14$ and $12.37$ and the benchmarks to base
level. What it does not repair is the first task, and order decides how much is left of
it: chain~A ends with Welsh at $40.80$ against $40.7$ for the base model, no trace of the
run that produced it, while chain~B ends with medicine at $10.91$, worse than the $8.67$
it reached but still better than the base model's $12.8$.
The model learns each new task as if nothing had happened before. It does not reliably keep the previous one.

Our own regularizer cannot prevent this: it points at the frozen base model (Equation~\ref{eq:kl}), which does not know Welsh. Learning-without-Forgetting anchors to the previous model and EWC accumulates its Fisher across tasks for this reason \citep{li2017lwf, kirkpatrick2017ewc}.

\paragraph{Two controlled variants of run~2.}
Each run restarts the learning rate at $10^{-4}$, so the collapse could come from a fresh
schedule rather than from stacking. Repeating run~2 of chain~A at $2\times10^{-5}$, one
fifth of the rate, learns the image style much less (CMMD $0.26$ against $0.13$) and
still degrades Welsh from $9.56$ to $669$, sixteen times worse than the base model that
never learned it: the schedule sets the size of the collapse and stacking causes it. The second variant follows from the objective. Setting $\lambda_{\text{txt}}=1$ in Equation~\ref{eq:kl} over WikiText (the training split; perplexity is measured on the test split), with everything else identical, removes the damage: general-text perplexity ends at $13.53$ against $229.8$
without it and $13.71$ for the base model, while the style is learned as before (CMMD
$0.134$ against $0.132$). The term is a KL to the frozen base model, not a cross-entropy on text, so it adds no new training signal: it only pulls the text head back toward where it already was. It is some $300$ times smaller than the image cross-entropy, and it is the only gradient the text head receives. Welsh shows what that means: it ends at $168.8$ against $4634$ without the anchor, still four times the $40.7$ of a model that never saw Welsh.
The anchor restores what the base knew and cannot restore what only the adapter knew.
The failure is therefore not that the recipe lacks a term. What a single run cannot reveal is that the unconstrained head matters at all; only composition makes it visible. Run~2 of each chain was repeated with a different data order and adapter initialization: the benchmark damage replicates within sampling noise (chain~B: MME $968$ and $984$, POPE $.819$ and $.822$, MMLU $.440$ and $.449$), while general-text perplexity, measured on a collapsed model, varies between runs (chain~A $230$ and $270$, chain~B $1950$ and $1679$) without changing regime.

\section{Conclusion}
We measured cross-modal forgetting under parameter-efficient post-training of a unified
multimodal model, choosing every target by how poorly the base model already handled it
so that preservation is tested against a real intervention. Across two directions, five
targets, LoRA ranks $2$ to $64$, and KL weights $\lambda\in[0,1.5]$, the capabilities we
did not train stayed at the base level within evaluation noise, including under a strong
shift toward a language the base barely knew, and no adapter degrades free-form English. Where a cost
appeared it was small and confined to grounded visual understanding at the strongest
settings. That stability belongs to a single run, not to the recipe: applied in sequence,
the same configuration destroys what it had just learned and pushes general text far
below a model that was never fine-tuned, on the very measures that stay flat after one
run, which is what makes their flatness informative. The damage follows from the objective, which regularizes only the modality being trained; the remedy is to anchor every head the shared weights feed, not only the one under training, which removes the damage to the untrained capabilities, though not the loss of the task learned one run earlier. Our probes
place an upper bound on how much of the new language reaches the visual pathway, and that
bound is close to zero. Repeated seeds, and a replication on a fully tokenized
early-fusion model, are the measurements we would add next.

\paragraph{Limitations.}
Our evidence comes from one unified model, Janus-Pro-7B, whose understanding and
generation encoders are separate modules over a shared backbone. LoRA acts on the backbone and nominally on the understanding encoder, but not on the VQ tokenizer or the image head, and the SigLIP adapters never receive gradient (\S\ref{sec:setup}), so in practice only the shared backbone is trained. Both modality-specific encoders are therefore untouched, and whatever damage we measure travels through the backbone alone. POPE and MME show small shifts, most visibly for the line-sketch target.
We do not vary the base model or its size. Whether our conclusions hold for fully tokenized
early-fusion models such as Chameleon, where one token stream shares all the weights, is
untested and is the central open question. We could not include one because their
released generation path is either disabled or reachable only through a bespoke stack.

Our central single-run claim is negative, and single-seed evidence cannot establish it.
The standard errors we report are sampling errors of the evaluation sets at fixed
weights. They say nothing about run-to-run variance. The four benchmarks were not chosen
for sensitivity to cross-modal damage. The chains of \S\ref{sec:chain} show that all four
move when damage is catastrophic, which is a limit from one side only: we show the
benchmarks are not inert, not how small a damage they could detect. A graded control
would supply that floor. The single-run results should therefore be read as a failure to
detect cross-modal forgetting under this protocol, not as proof that it is absent. Our
sequential protocol is short: three runs in two orders, with the middle task held fixed
rather than varied and only its second run repeated at a second initialization. It does
not show how the damage grows with sequence length, nor separate the middle task's
modality from the effect of stacking. 
We do not test instruction following, which the text adapters visibly change, and our chains re-train one adapter throughout: merging each run into the base, or starting from a random nonzero adapter, would separate adapter reuse from weight displacement.
Of the anti-forgetting objectives we cover only
regularization toward a frozen reference, not replay, orthogonality-constrained methods
such as O-LoRA \citep{wang2023olora}, or routing. Free-form text quality is assessed by
distribution metrics, not a human study.

\section*{Acknowledgements}
This publication is part of project 2026.008 of the Computing Time National Computing
Facilities program which is financed by the Dutch Research Council (NWO) under the grant
\url{https://doi.org/10.61686/WBFDE48237}.

\bibliographystyle{plainnat}
\bibliography{references}
\appendix

\section{Implementation details}
\label{app:impl}

\paragraph{Model.}
All experiments use Janus-Pro-7B (\texttt{deepseek-community/Janus-Pro-7B}), loaded in
\texttt{bfloat16} through the Hugging Face \texttt{JanusForConditionalGeneration} class.
Janus-Pro is a unified autoregressive model with one shared transformer backbone and two
\emph{separate} modality encoders. The SigLIP vision encoder handles understanding
(image$\to$text) and the vector-quantized (VQ) tokenizer with an image head handles
generation (text$\to$image). An image is a $24\times24=576$ grid of discrete tokens
drawn from a codebook of size $16384$. We use the two directions through the processor's
text mode for language modeling and image mode (\texttt{max\_new\_tokens}$=576$, then
VQ decoding). LoRA is applied to the shared backbone and, because targets are matched by
module-name suffix, to the $q,k,v$ projections of the SigLIP encoder ($24$ layers, $72$
modules). SigLIP's output projection is named \texttt{out\_proj} and is therefore not
matched. These $72$ adapters are never updated: no training in either direction encodes an image for understanding, and their $B$ matrices are exactly zero in every saved adapter. The VQ tokenizer and image head are frozen. Appendix~\ref{app:dw} reports
backbone modules only and says so.

\paragraph{LoRA configuration (fixed across all runs).}
Every adapter targets the seven projections of each backbone layer,
\texttt{q,k,v,o\_proj} for attention and \texttt{gate,up,down\_proj} for MLP. We set
$\alpha=2r$ so the LoRA scaling $\alpha/r=2$ is constant across the rank sweep. We use
\texttt{bias=none}, the causal-LM task type, and no rsLoRA. The resulting weight change
is $\Delta W=(\alpha/r)BA$. Within a sweep only the swept variable changes, rank $r$ or
KL weight $\lambda$.

\paragraph{Optimization (common).}
We use AdamW with $\beta=(0.9,0.95)$ (matching the Janus pre-training $\beta_2$) and no
weight decay. We train with a cosine schedule with $5\%$ warmup, gradient accumulation
$8$, gradient clipping $1.0$, \texttt{bfloat16} on a single H100. With the regularizer
the loss is
$\mathcal{L}=\mathrm{CE}+\lambda\,\mathrm{KL}(p_{\text{adapter}}\|p_{\text{base}})$.
$p_{\text{base}}$ comes from a \texttt{disable\_adapter()} forward pass without
gradients on the same inputs ($\approx$$2\times$ forward cost).

\paragraph{Uncertainty quantification.}
Every number is reported as mean $\pm$ one standard error, computed without retraining.
The accuracy benchmarks (MMLU, POPE, MME) are proportions over a fixed evaluation set, so
we use the Wilson form \citep{wilson1927,brown2001interval}. Per-instance metrics (DirSim,
DINO, CLIP-IQA) use $s/\sqrt{n}$. CMMD and KID have no closed-form variance and use a
nonparametric bootstrap over both sets ($B{=}2000$)
\citep{efron1993bootstrap,binkowski2018demystifying}. Perplexity propagates the error
through the exponential. All of these quantify evaluation-set sampling noise at fixed
weights, not training-seed variance, which is why the small benchmark movements of
\S\ref{sec:boundary} are read cautiously.

\begin{table}[htbp]
\caption{\textbf{The two evaluation configurations}, given as $n$ with the base model's
score in parentheses. The full sets serve the single-run results and the image sweeps; the
subsample serves the text sweeps and the sequential runs of \S\ref{sec:chain}. It is drawn
once and reused, so runs within a table are comparable. Numbers are never compared across
the two configurations, and every table is labeled \emph{Base (full)} or \emph{Base
(subsampled)}. WikiText perplexity uses the same test set in both, where the base model
scores $13.71$.}
\label{tab:evalsizes}
\centering\small
\setlength{\tabcolsep}{8pt}
\begin{tabular}{lccc}
\toprule
Configuration & MMLU & POPE & MME \\
\midrule
Full       & $1425$ ($.492$) & $3000$ ($.877$) & $2374$ ($1785$) \\
Subsampled & $570$  ($.481$) & $1000$ ($.875$) & $600$  ($1130$) \\
\bottomrule
\end{tabular}
\end{table}

\paragraph{Text (poetry) runs.}
Data: \texttt{biglam/gutenberg-poetry-corpus}, grouped by source document with
front-matter removed and windowed to $512$ tokens ($6000$ train / $200$ val windows).
Fixed: $r{=}16$, $\alpha{=}32$, dropout $0.05$, LR $10^{-4}$, $3$ epochs ($150$-min
wall-clock budget). Rank sweep: $r\in\{4,8,16,32,64\}$ at $\lambda{=}0.5$; $\lambda$
sweep: $\lambda\in\{0,0.25,0.5,1,1.5\}$ at $r{=}16$.

\paragraph{Image (Naruto) runs.}
Data is \texttt{lambdalabs/naruto-blip-captions} ($\approx$$1141$ image--caption pairs).
Acquisition uses $80$ held-out prompts. Final configuration (Table~\ref{tab:panel}) uses
$r{=}48$, $\alpha{=}96$, dropout $0.10$, LR $10^{-4}$, $6$ epochs, no caption prefix,
$\lambda{=}0.5$. Rank sweep varies $r\in\{2,\dots,64\}$ at $\lambda{=}0.5$. $\lambda$
sweep is at $r{=}32$.

\paragraph{Medicine, Welsh and sketch runs.}
Medicine uses \texttt{ccdv/pubmed-summarization} with $r{=}32$, $\alpha{=}64$,
$\lambda{=}0.25$. Welsh uses \texttt{wikimedia/wikipedia} (Welsh subset) with $r{=}32$,
$\alpha{=}64$, $\lambda{=}0.5$, dropout $0.05$, LR $10^{-4}$, $3$ epochs, gradient
accumulation $8$, cosine schedule. The data is $40{,}000$ documents subsampled to
$6{,}000$ training and $200$ validation windows.

\paragraph{Prompts and image generation.}
Janus generates images token by token from a natural-language prompt (text$\to$image).
Every generated image in this paper is therefore conditioned on an explicit text prompt.
During training the conditioning text is the dataset's BLIP caption. The
\textit{Capacity} and \textit{+KL} configurations prepend a fixed style trigger (``anime
sprite in Naruto style, ''). \textit{Scaled} and \textit{Final} drop the trigger. The
Final adapter therefore applies the style \emph{unconditionally}, with no trigger token
at inference. The style metrics (Table~\ref{tab:panel}) are computed on images generated
from $80$ held-out Naruto captions. The qualitative grids use a fixed, curated set of
generic prompts that mix in-domain and clearly out-of-domain subjects (e.g.\ ``a young
man with blonde spiky hair'', ``a samurai warrior holding a katana'', ``an old wizard
with a long white beard''). These prompts test whether the style carries over to
arbitrary content. The exact prompt is printed under each row of every grid
(Figures~\ref{fig:janus_qualitative_comparison} and \ref{fig:kl_visual}). The
cross-modal check under the text adapter (Figure~\ref{fig:t2i_welsh}) uses three neutral
prompts: an animal, an architectural scene and a surreal scene. All conditions share the
same seed schedule, so within a row the adapter is the only thing that varies.

\section{Target selection and how much each benchmark can lose}
\label{app:targets}
\subsection{Selecting the language target}
\label{app:langprobe}
Before training, we scored candidate language targets by $\times$control. This is the
base model's perplexity on the candidate corpus divided by its perplexity on an English
WikiText control. Control perplexity was $14.22$ under this probe's chunking. The
candidates were real, available corpora: low-resource natural languages from Wikipedia,
plus a synthetic substitution-cipher language built from English as the most distant
possible reference point. Table~\ref{tab:langprobe} reports the result. Poetry and medicine were scored in one probe run and Welsh in another, with the same construction and the same control corpus.

We did not take the most distant candidate. The cipher is by far the most distant
($10.3\times$), but it is a fixed character-by-character remapping of English, and an
adapter could fit it by learning the substitution table. That would make the
``learning'' a shortcut rather than a new linguistic ability. Klingon is slightly more
distant than Welsh ($2.51\times$ vs.\ $2.34\times$). Its available corpus is a few
thousand short sentences, too small to separate learning from memorization at our
training budget. Welsh combines a real distance from the base model with a large
naturally occurring corpus, which is why we chose it. The absolute perplexities in this
table are not comparable with Table~\ref{tab:targets}. This probe scores whole articles
under its own chunking against a $14.22$ control. The training evaluation uses the
$512$-token windows of Appendix~\ref{app:impl} and gives a base Welsh perplexity of
$40.7$. Only the $\times$control ratio is meant to carry across.

\begin{table}[htbp]\centering\small
\caption{Language-target probe on the base model only, before any training. $\times$control is the base model's perplexity on the candidate divided by its perplexity on an English control ($14.22$ under this probe's chunking); higher means the corpus is further from what the base model already handles well. Welsh was chosen over the two more distant candidates for the reasons given in the text.}
\setlength{\tabcolsep}{8pt}
\begin{tabular}{lccl}
\toprule
Candidate & Base PPL & $\times$control & Note \\
\midrule
Cipher (substitution) & $146.8$ & $10.33$ & deterministic remapping of English \\
Klingon               & $35.7$  & $2.51$  & constructed; corpus too small \\
\textbf{Welsh}        & $33.3$  & $\mathbf{2.34}$ & \textbf{selected}: distant and abundant \\
Yoruba                & $24.0$  & $1.69$  & closer to the base \\
\bottomrule
\end{tabular}
\label{tab:langprobe}
\end{table}

\subsection{How much each benchmark can lose}
\label{app:headroom}
Table~\ref{tab:headroom} shows the correction from Section~\ref{sec:setup} applied to
the base model for both evaluation configurations. The full configuration uses
$n{=}1425$ for MMLU, $3000$ for POPE and $2374$ for MME. The subsampled configuration
uses $570$, $1000$ and $600$. Intervals are Wilson intervals computed on the observed accuracy, with both endpoints then mapped through the correction, which is monotone increasing.
The correction assumes the guessed items are balanced between the answers; POPE's
yes-ratio stays between $0.44$ and $0.49$ in every run we report, so answer bias does
not distort these estimates. It does not apply to perplexity.

\begin{table}[htbp]\centering\small
\caption{How much of the base model's accuracy is real knowledge that could be lost. MME is corrected on its item accuracy, not on the composite score.}
\label{tab:headroom}
\begin{tabular}{llccccc}
\toprule
Config & Benchmark & $k$ & $n$ & accuracy & chance & \textbf{headroom} (95\% CI) \\
\midrule
Full       & MMLU & 4 & 1425 & .492 & .169 & \textbf{.323} \; [.288, .357] \\
Full       & POPE & 2 & 3000 & .877 & .123 & \textbf{.755} \; [.730, .777] \\
Full       & MME  & 2 & 2374 & .788 & .212 & \textbf{.576} \; [.542, .608] \\
\midrule
Subsampled & MMLU & 4 &  570 & .481 & .173 & \textbf{.308} \; [.253, .362] \\
Subsampled & POPE & 2 & 1000 & .875 & .125 & \textbf{.750} \; [.706, .788] \\
Subsampled & MME  & 2 &  600 & .803 & .197 & \textbf{.607} \; [.539, .666] \\
\bottomrule
\end{tabular}
\end{table}

Applying the same correction to all $30$ fine-tuned runs leaves every conclusion of
Section~\ref{sec:gen} and Section~\ref{sec:text} intact. No adapter's corrected
knowledge on any benchmark falls outside the base model's confidence interval for its
configuration, even though the correction amplifies differences by $k/(k-1)$. MMLU is
the weakest of our three multiple-choice measures in this respect, with only about a
third of its observed accuracy attributable to knowledge. POPE is the strongest.

\section{Image-generation configuration}
\label{app:imgconfig}

\subsection{Incremental ablation}
\label{app:ablation}
We build the Final configuration by adding components a few at a time
(Table~\ref{tab:panel}). The naive start, attention-only LoRA at rank 8 with a fixed
caption-prefix trigger, degraded the base model and is omitted. \emph{Capacity} restores
trainable capacity (rank 16, attention and MLP); \emph{+KL} adds the regularizer
($\lambda{=}0.5$); \emph{Scaled} drops the training-time trigger and raises rank, learning
rate and dropout (rank 32, $8\times10^{-5}$, dropout 0.1, six epochs); \emph{Final} raises
the rank to 48 and the learning rate to $10^{-4}$. The \emph{Joint} conditions additionally
alternate a $50/50$ domain-controlled text stream into the same adapter (matched: Naruto
franchise; far: physics and mathematics). The corresponding images are in
Figure~\ref{fig:janus_qualitative_comparison}.

\begin{table}[htbp]
\centering\small
\setlength{\tabcolsep}{5pt}
\caption{Building the image-generation configuration step by step; each column adds components to the previous one, as described in the text. The naive start (attention-only, rank 8, prefix-gated) degraded the base model and is omitted. CLIP DirSim turning positive at \textit{Scaled} marks the point where real style transfer begins.}
\renewcommand{\arraystretch}{1.05}
\begin{tabular}{lcccccc}
\toprule
 & \textbf{Capacity} & \textbf{+KL} & \textbf{Scaled} & \textbf{Final} & \textbf{Joint} & \textbf{Joint} \\
 & & & & & \textbf{(matched)} & \textbf{(far)} \\
\midrule
\multicolumn{7}{l}{\textit{Configuration (bold = changed vs.\ previous column)}}\\
Rank $r$              & 16 & 16 & \textbf{32} & \textbf{48} & 48 & 48 \\
LR ($\times10^{-5}$)  & 5 & 5 & \textbf{8} & \textbf{10} & 10 & 10 \\
Epochs               & 8 & 8 & \textbf{6} & 6 & 6 & 6 \\
LoRA dropout         & .05 & .05 & \textbf{.10} & .10 & .10 & .10 \\
Train prefix         & yes & yes & \textbf{no} & no & no & no \\
KL $\lambda$         & 0 & \textbf{0.5} & 0.5 & 0.5 & 0.5 & 0.5 \\
Joint text           & --- & --- & --- & --- & \textbf{matched} & \textbf{far} \\
\midrule
\multicolumn{7}{l}{\textit{Acquisition (vs.\ real style, $N{=}80$; bold = best)}}\\
KID $\downarrow$       & $.039$ & $.017$ & $.016$ & $.016$ & $.016$ & $\mathbf{.014}$ \\
CMMD $\downarrow$      & $.189$ & $\mathbf{.105}$ & $.120$ & $.110$ & $.131$ & $.134$ \\
DINO sim $\uparrow$    & $.629$ & $\mathbf{.663}$ & $.624$ & $.640$ & $.633$ & $.637$ \\
CLIP-IQA $\uparrow$    & $.659$ & $\mathbf{.750}$ & $.694$ & $.701$ & $.725$ & $.704$ \\
CLIP DirSim $\uparrow$ & $-.059$ & $-.033$ & $.157$ & $\mathbf{.163}$ & $.146$ & $.151$ \\
\bottomrule
\end{tabular}
\label{tab:panel}
\end{table}

\paragraph{The KL weight has a best value for learning the style.}
Table~\ref{tab:klimg} varies $\lambda$ at rank 32 for image generation. Unlike text, the
learned style peaks at an intermediate $\lambda$ and is best near $\lambda{=}0.25$ to
$0.5$. Performance collapses for $\lambda\!\geq\!1$, where the adapter is held too close
to the base model. $\lambda{=}0$ already works, so for images the KL penalty refines the
recipe instead of making it possible. Retention stays at the base level across the whole
$\lambda$ range, so the penalty changes what is learned with no measurable cost to the
other capabilities. The CLIP style metrics put the optimum at $\lambda{=}0.25$, yet we
use $\lambda{=}0.5$. Re-scoring the same generated images against held-out and training
reference images rules out noise and memorization: the $\lambda$-ranking by CMMD is
identical for the two sets, and nearest-neighbor DINOv2 similarity to training images
is lower than to held-out images at every $\lambda$. What remains is a difference
between metrics. CLIP rewards closeness to the style distribution alone, whereas a
distribution distance in DINOv2 space and a content-preservation measure both favor
$\lambda{=}0.5$ ($.035$ against $.041$, and $.39$ against $.32$), as does an independent
VLM judge (Qwen2-VL-7B, order-consistent pairwise preferences on the same $80$ prompts,
decisive win rate $76\%$, $95\%$ CI $[55,89]$).

\begin{table}[htbp]
\centering\small
\caption{Image-generation $\lambda$ sweep at rank $32$. The learned style (bold = best) peaks near $\lambda{=}0.25$--$0.5$ and collapses for $\lambda\!\geq\!1$; retention stays at the base level throughout, so the KL penalty changes what is learned at no measurable cost to the other capabilities.}
\setlength{\tabcolsep}{6pt}
\begin{tabular}{lcccccc}
\toprule
Metric & Base & $\lambda{=}0$ & $\lambda{=}0.25$ & $\lambda{=}0.5$ & $\lambda{=}1.0$ & $\lambda{=}1.5$ \\
\midrule
\multicolumn{7}{l}{\textit{Acquisition (vs.\ real style images)}}\\
CMMD $\downarrow$      & --- & $.155$ & $\mathbf{.100}$ & $.108$ & $.205$ & $.221$ \\
DINO sim $\uparrow$    & --- & $.626$ & $\mathbf{.647}$ & $.643$ & $.439$ & $.388$ \\
CLIP DirSim $\uparrow$ & --- & $.157$ & $\mathbf{.164}$ & $.152$ & $.049$ & $.024$ \\
CLIP-IQA $\uparrow$    & --- & $.666$ & $.691$ & $.691$ & $.738$ & $\mathbf{.739}$ \\
\midrule
\multicolumn{7}{l}{\textit{Untrained capabilities}}\\
WikiText PPL $\downarrow$ \tiny($\pm.21$) & $13.71$ & $13.67$ & $13.68$ & $13.78$ & $13.76$ & $13.76$ \\
MMLU $\uparrow$ \tiny($\pm.013$)          & $.492$ & $.495$ & $.493$ & $.497$ & $.493$ & $.492$ \\
POPE acc $\uparrow$ \tiny($\pm.006$)      & $.877$ & $.879$ & $.877$ & $.879$ & $.877$ & $.873$ \\
MME $\uparrow$ \tiny($\pm31$)             & $1785$ & $1783$ & $1776$ & $1796$ & $1802$ & $1810$ \\
\bottomrule
\end{tabular}
\label{tab:klimg}
\end{table}

\begin{figure*}[htbp]
\centering
\setlength{\tabcolsep}{1.5pt}\small
\begin{tabular}{cccccc}
\textbf{Base} & \textbf{$\lambda{=}0$} & \textbf{$\lambda{=}0.25$} & \textbf{$\lambda{=}0.5$} & \textbf{$\lambda{=}1.0$} & \textbf{$\lambda{=}1.5$} \\
\midrule
\includegraphics[width=0.155\textwidth]{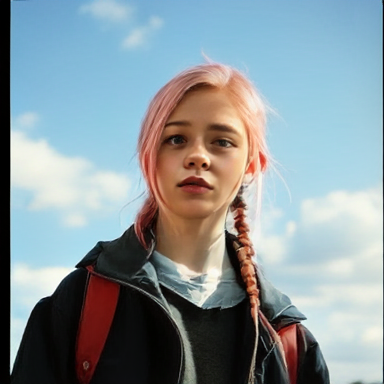} &
\includegraphics[width=0.155\textwidth]{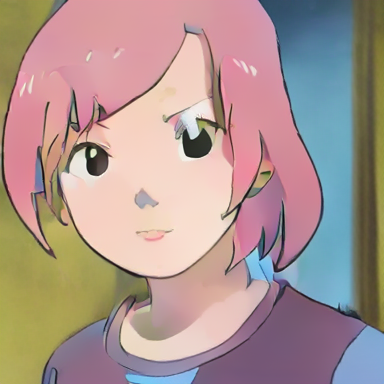} &
\includegraphics[width=0.155\textwidth]{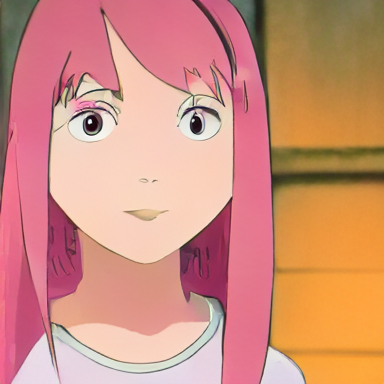} &
\includegraphics[width=0.155\textwidth]{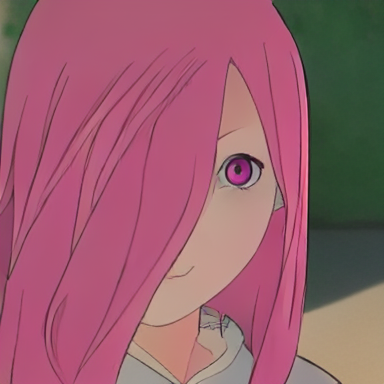} &
\includegraphics[width=0.155\textwidth]{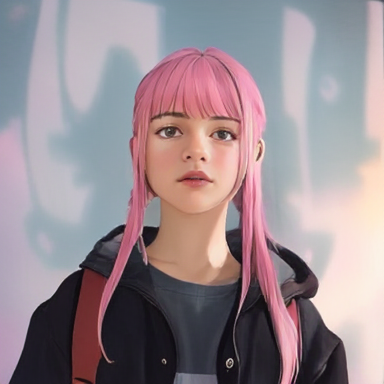} &
\includegraphics[width=0.155\textwidth]{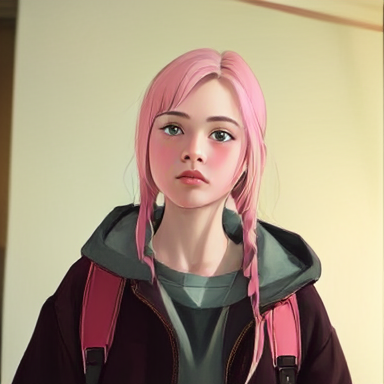} \\
\multicolumn{6}{l}{\footnotesize\textit{Prompt:} a teenage girl with pink hair} \\
\addlinespace[2pt]
\includegraphics[width=0.155\textwidth]{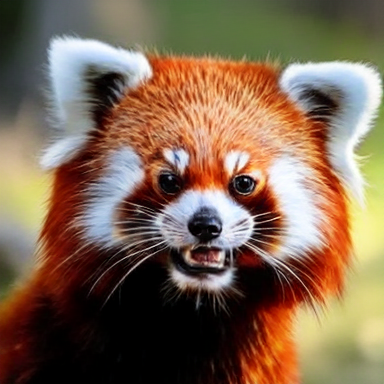} &
\includegraphics[width=0.155\textwidth]{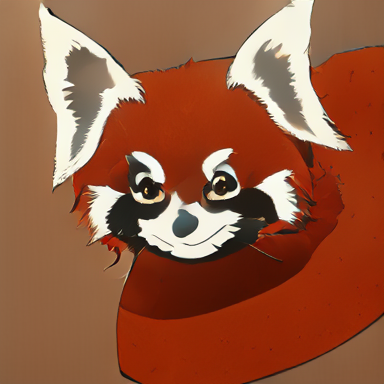} &
\includegraphics[width=0.155\textwidth]{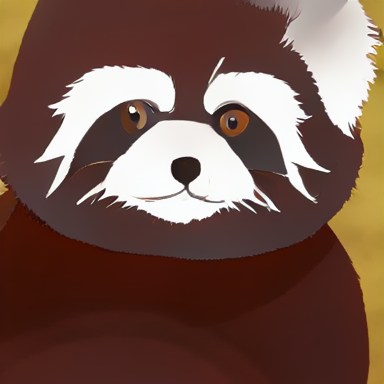} &
\includegraphics[width=0.155\textwidth]{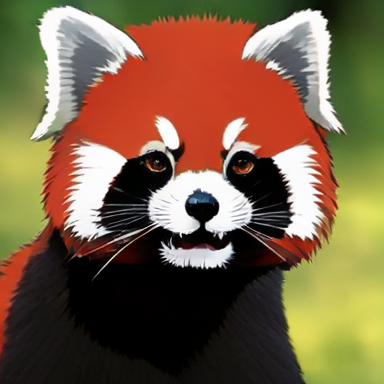} &
\includegraphics[width=0.155\textwidth]{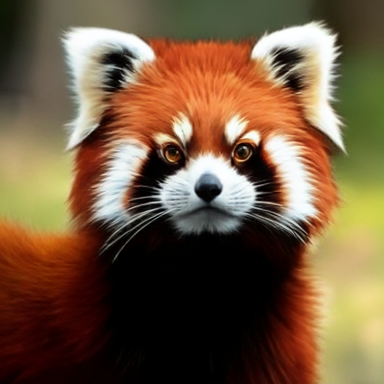} &
\includegraphics[width=0.155\textwidth]{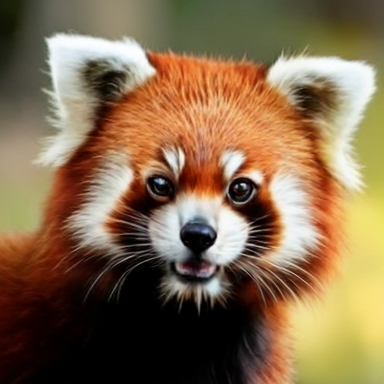} \\
\multicolumn{6}{l}{\footnotesize\textit{Prompt:} a red panda with a black face and brown eyes} \\
\addlinespace[2pt]
\includegraphics[width=0.155\textwidth]{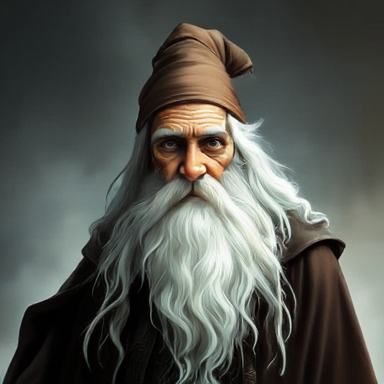} &
\includegraphics[width=0.155\textwidth]{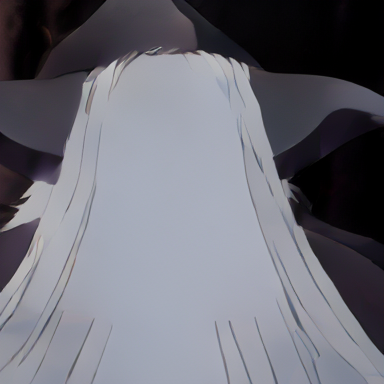} &
\includegraphics[width=0.155\textwidth]{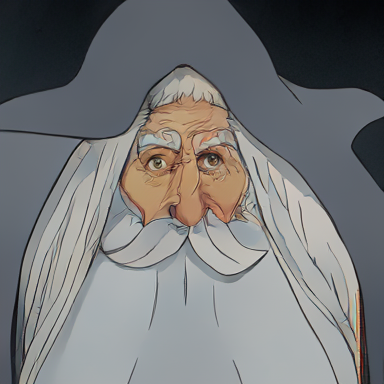} &
\includegraphics[width=0.155\textwidth]{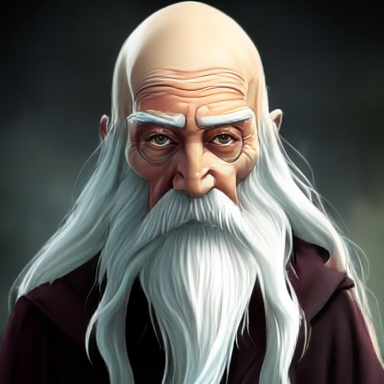} &
\includegraphics[width=0.155\textwidth]{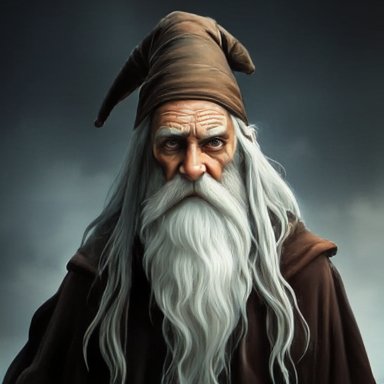} &
\includegraphics[width=0.155\textwidth]{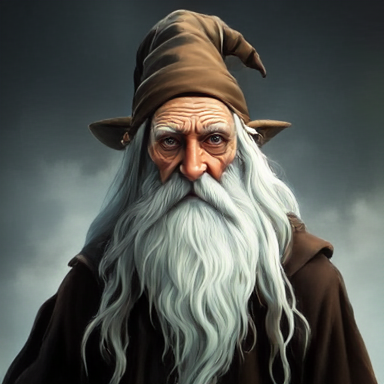} \\
\multicolumn{6}{l}{\footnotesize\textit{Prompt:} an old wizard with a long white beard} \\
\end{tabular}
\caption{Effect of the KL weight $\lambda$ on the learned visual style (rank $32$), on the same three prompts and seeds as Figure~\ref{fig:janus_qualitative_comparison}. Style transfer is strong for $\lambda\!\in\![0,0.5]$ and visibly collapses back toward the base model for $\lambda\!\geq\!1$, the visual counterpart of the DirSim column in Table~\ref{tab:klimg}.}
\label{fig:kl_visual}
\end{figure*}

\begin{figure*}[htbp]
\centering
\setlength{\tabcolsep}{1.5pt}
\renewcommand{\arraystretch}{1.0}
\small
\begin{tabular}{ccccccc}
\textbf{No adapter} & \textbf{Capacity} & \textbf{+KL} & \textbf{Scaled} & \textbf{Final} & \textbf{Joint (matched)} & \textbf{Joint (far)} \\
\textit{base 7B} & \textit{r16, +MLP} & \textit{+ KL reg} & \textit{r32, untriggered} & \textit{r48, LR $10^{-4}$} & \textit{+ Naruto wiki} & \textit{+ physics wiki} \\
\midrule
\includegraphics[width=0.135\textwidth]{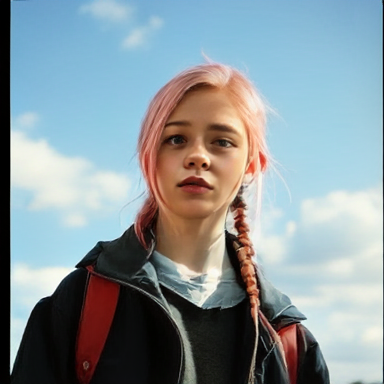} &
\includegraphics[width=0.135\textwidth]{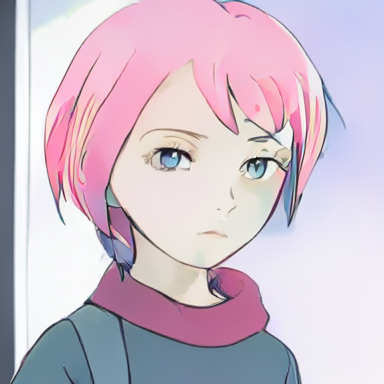} &
\includegraphics[width=0.135\textwidth]{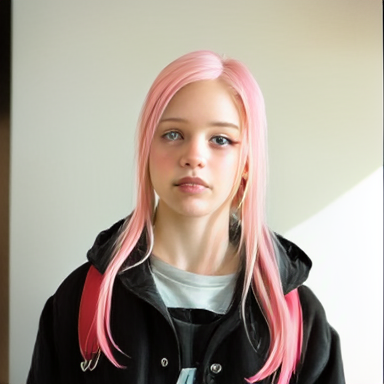} &
\includegraphics[width=0.135\textwidth]{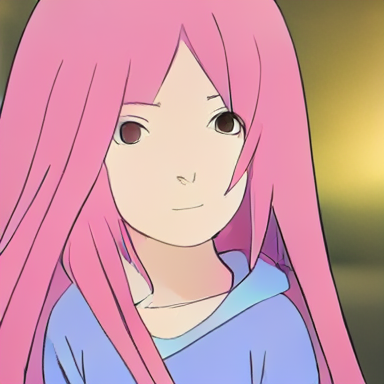} &
\includegraphics[width=0.135\textwidth]{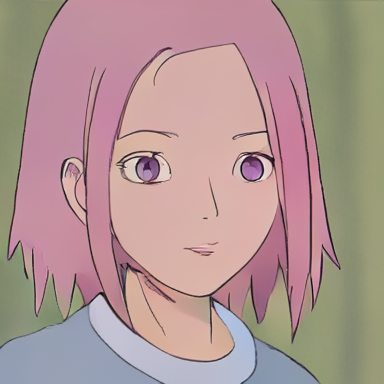} &
\includegraphics[width=0.135\textwidth]{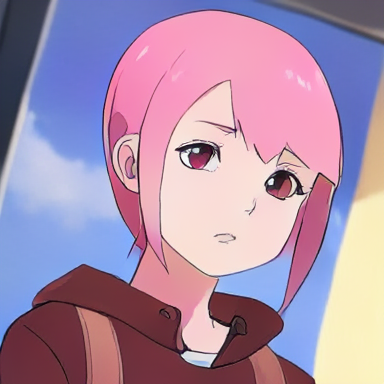} &
\includegraphics[width=0.135\textwidth]{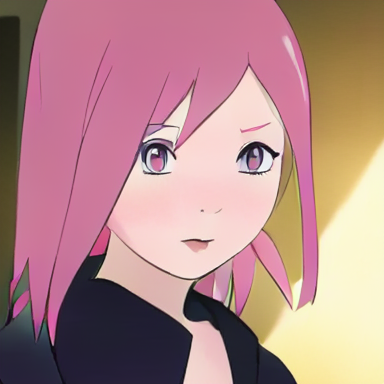} \\
\multicolumn{7}{l}{\footnotesize\textit{Prompt:} a teenage girl with pink hair} \\
\addlinespace[2pt]
\includegraphics[width=0.135\textwidth]{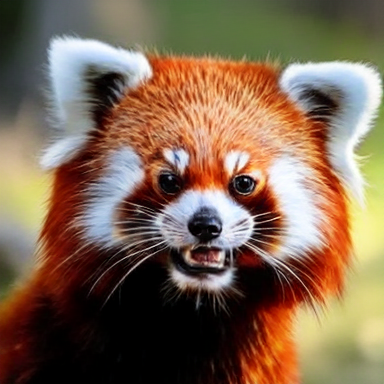} &
\includegraphics[width=0.135\textwidth]{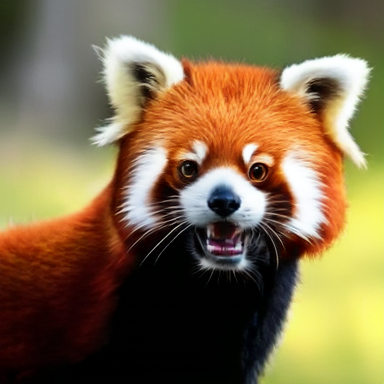} &
\includegraphics[width=0.135\textwidth]{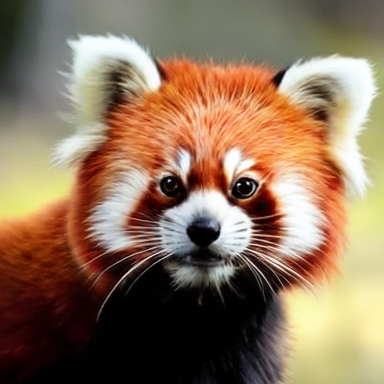} &
\includegraphics[width=0.135\textwidth]{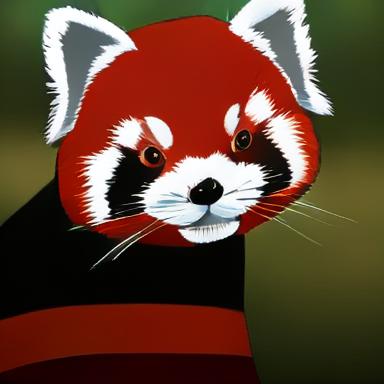} &
\includegraphics[width=0.135\textwidth]{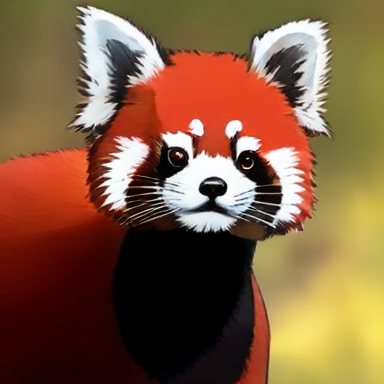} &
\includegraphics[width=0.135\textwidth]{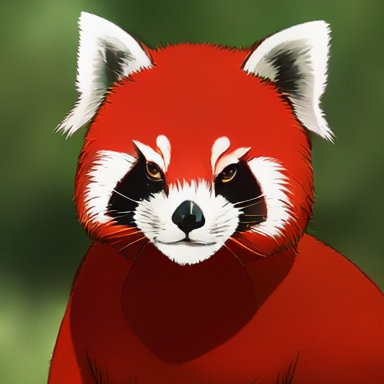} &
\includegraphics[width=0.135\textwidth]{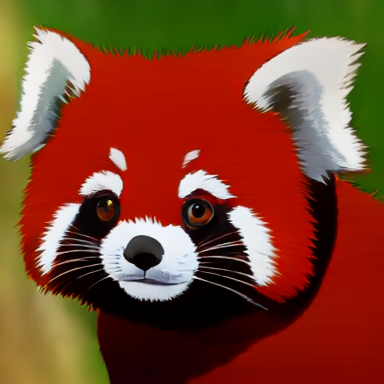} \\
\multicolumn{7}{l}{\footnotesize\textit{Prompt:} a red panda with a black face and brown eyes} \\
\addlinespace[2pt]
\includegraphics[width=0.135\textwidth]{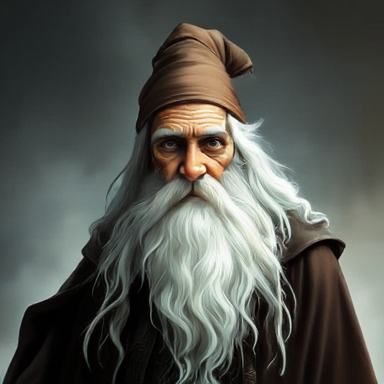} &
\includegraphics[width=0.135\textwidth]{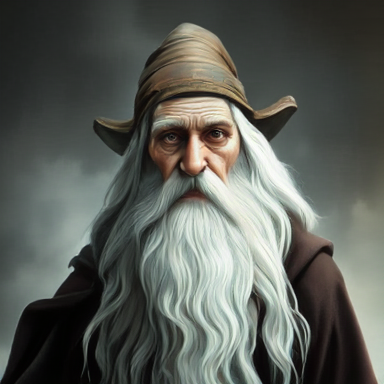} &
\includegraphics[width=0.135\textwidth]{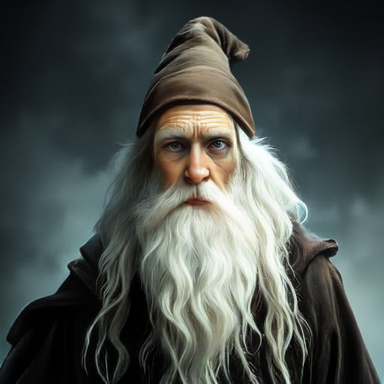} &
\includegraphics[width=0.135\textwidth]{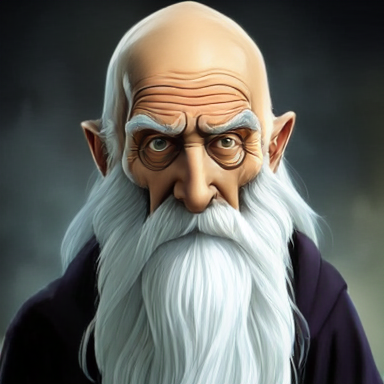} &
\includegraphics[width=0.135\textwidth]{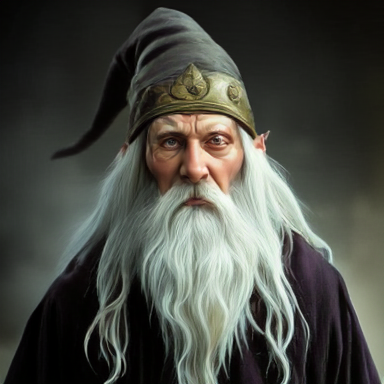} &
\includegraphics[width=0.135\textwidth]{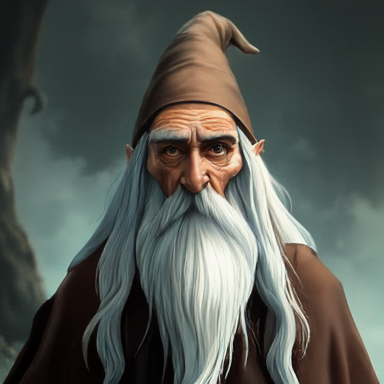} &
\includegraphics[width=0.135\textwidth]{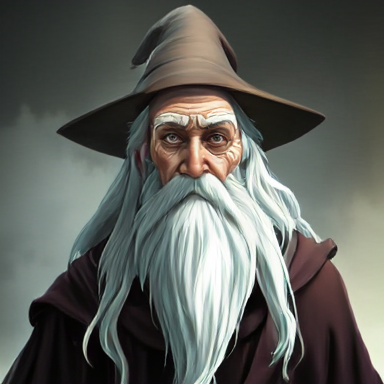} \\
\multicolumn{7}{l}{\footnotesize\textit{Prompt:} an old wizard with a long white beard} \\
\end{tabular}
\caption{The seven training conditions on the same prompts and seeds, so within a row the adapter is the only thing that changes. Columns 1--5 follow the image-only LoRA from the naive start to \textit{Final}; columns 6--7 add a joint text stream from two distant Wikipedia domains. The first prompt is in the style's domain, the other two deliberately are not.}
\label{fig:janus_qualitative_comparison}
\end{figure*}

\paragraph{Joint training: a corpus-size effect on perplexity.}
The \emph{Joint} conditions are the sensitivity control moved to the main text
(Table~\ref{tab:joint}): held-out perplexity inflates only when the text corpus is small.
The controlled far@320 run uses a distant domain with a matched chunk budget and reaches
$55.1$, higher than the near-domain matched run, while MMLU and POPE stay flat. The
inflation tracks corpus size, not domain distance.

\begin{table}[htbp]
\caption{\textbf{Sensitivity control.} A text stream alternated into the image adapter's training at three corpus sizes. Held-out perplexity moves by tens of standard errors when the text corpus is small, tracking corpus size rather than domain distance; knowledge and understanding are unaffected. Full configuration, single seed.}
\label{tab:joint}
\centering\small
\setlength{\tabcolsep}{6pt}
\begin{tabular}{llcccc}
\toprule
\textbf{Condition} & \textbf{Text corpus} & \textbf{PPL}$\downarrow$ & \textbf{MMLU}$\uparrow$ & \textbf{POPE acc}$\uparrow$ & \textbf{MME}$\uparrow$ \\
 & & \tiny($\pm.21$) & \tiny($\pm.013$) & \tiny($\pm.006$) & \tiny($\pm31$) \\
\midrule
Base (no adapter) & ---                  & $13.71$ & $0.492$ & $0.877$ & $1785$ \\
Final (image-only)& ---                  & $13.80$ & $0.491$ & $0.879$ & $1789$ \\
Joint matched     & near, 320 chunks     & $47.87$ & $0.505$ & $0.876$ & $1747$ \\
Joint far         & distant, 7672 chunks & $12.46$ & $0.491$ & $0.878$ & $1795$ \\
Joint far@320     & distant, 320 chunks  & $55.10$ & $0.481$ & $0.873$ & $1736$ \\
\bottomrule
\end{tabular}
\end{table}

\subsection{Rank sweep for image generation}
\label{app:rank}

\begin{table}[htbp]
\caption{\textbf{Fine-tuning image generation: rank sweep.} How well the style is learned (bold = best) improves and then levels off; every untrained capability stays at the base level at all ranks. The base model has no DirSim by definition, and its CMMD was not computed in this sweep. Standard errors depend on the evaluation set, not on the adapter, so they are given once per row.}
\label{tab:rank}
\centering\small
\begin{tabular}{lccccccc}
\toprule
& Base (full) & $r{=}2$ & $r{=}4$ & $r{=}8$ & $r{=}16$ & $r{=}32$ & $r{=}64$ \\
\midrule
\multicolumn{8}{l}{\emph{Acquisition} (vs.\ real style images)}\\
CMMD $\downarrow$        & n/a & .139 & .124 & .126 & .105 & .108 & \textbf{.097} \\
CLIP DirSim $\uparrow$   & n/a & .143 & .142 & .143 & .154 & .153 & \textbf{.158} \\
\midrule
\multicolumn{8}{l}{\emph{Untrained capabilities}}\\
WikiText PPL $\downarrow$ \tiny($\pm.21$) & 13.71 & 13.66 & 13.70 & 13.76 & 13.74 & 13.78 & 13.76 \\
MMLU $\uparrow$ \tiny($\pm.013$)          & .492 & .491 & .491 & .491 & .501 & .497 & .489 \\
POPE acc $\uparrow$ \tiny($\pm.006$)      & .877 & .879 & .881 & .878 & .878 & .879 & .878 \\
MME $\uparrow$ \tiny($\pm31$)             & 1785 & 1777 & 1777 & 1786 & 1786 & 1796 & 1781 \\
\bottomrule
\end{tabular}
\end{table}

\subsection{Generalization to a second visual style}
\label{app:sketch}
To check that the image-generation result is not specific to one style, we repeat the
image protocol on a second, unrelated dataset. We use black-and-white line-sketch scenes
(\texttt{zoheb/sketch-scene}) with the \emph{same} configuration as for Naruto
($r{=}48$, $\alpha{=}96$, $\lambda{=}0.5$, LR $10^{-4}$). The main result holds
(Table~\ref{tab:sketch}). The adapter learns the style at the distribution level (CMMD
$0.253\!\to\!0.159$, KID $0.263\!\to\!0.096$, DINO similarity to the sketch reference
$0.133\!\to\!0.371$). Language and knowledge probes show no catastrophic forgetting
(WikiText PPL $13.71\!\to\!14.18$, MMLU $0.492\!\to\!0.479$).

\begin{figure}[htbp]
\centering
\begin{subfigure}{0.24\textwidth}
\includegraphics[width=\linewidth]{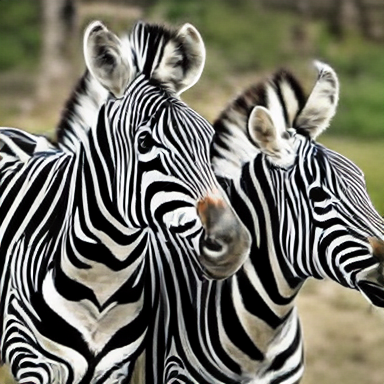}
\caption*{\footnotesize Base}
\end{subfigure}
\begin{subfigure}{0.24\textwidth}
\includegraphics[width=\linewidth]{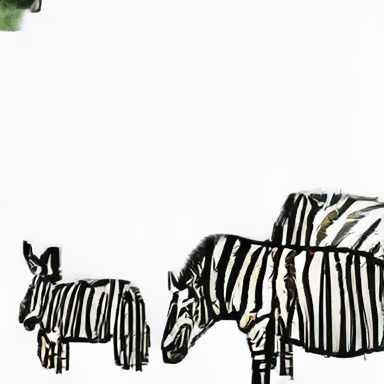}
\caption*{\footnotesize Sketch LoRA}
\end{subfigure}
\begin{subfigure}{0.24\textwidth}
\includegraphics[width=\linewidth]{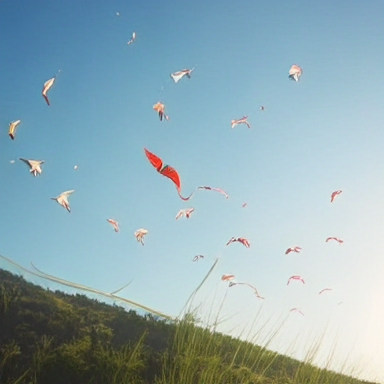}
\caption*{\footnotesize Base}
\end{subfigure}
\begin{subfigure}{0.24\textwidth}
\includegraphics[width=\linewidth]{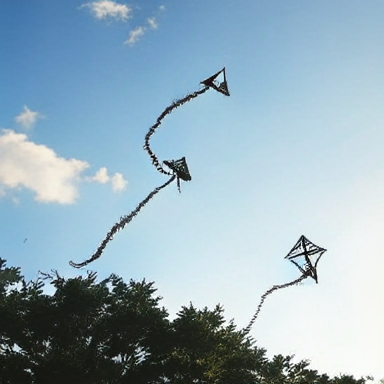}
\caption*{\footnotesize Sketch LoRA}
\end{subfigure}
\caption{Line-sketch scenes, same prompt and seed, base model vs.\ adapter. The style carries over for some prompts (left: white canvas, ink linework) but only partly for others (right: sketch strokes appear but photographic color remains).}
\label{fig:sketch}
\end{figure}

The replication gives two findings the main text uses. First, \emph{the PEFT
configuration depends on the dataset}: the recipe that fully learns a high-frequency,
color-saturated style only partly learns the sparse, low-frequency sketch style
(Figure~\ref{fig:sketch}). Per-prompt edit strength is about half (CLIP DirSim $0.086$
against $\approx\!0.15$ for Naruto at the same $\lambda$) and image quality drops
measurably (CLIP-IQA $0.747\!\to\!0.592$). Rank $64$ with $\lambda{=}0.25$ brings the
distribution closer (KID $0.096\!\to\!0.065$) without raising edit strength (DirSim
$0.086\!\to\!0.089$), so the limit looks intrinsic to the style rather than a matter of
capacity. Second, this is the target with the largest retention footprint of all five,
most visible on MME ($1785\!\to\!1713$, versus flat for Naruto), the third of the three
convergent settings of \S\ref{sec:boundary}.

\begin{table}[htbp]
\centering\small
\caption{The same configuration as for Naruto applied to line-sketch scenes (\texttt{zoheb/sketch-scene}). The style is learned at the distribution level and no probe shows catastrophic forgetting, but the per-prompt edit strength is about half of Naruto's. FID/ArtFID are omitted because they are numerically unstable at $n{=}80$.}
\setlength{\tabcolsep}{8pt}
\begin{tabular}{lcc}
\toprule
Metric & Base & Sketch LoRA \\
\midrule
\multicolumn{3}{l}{\textit{Acquisition (vs.\ sketch reference, $n{=}80$)}}\\
CMMD $\downarrow$      & $.253$ & $\mathbf{.159}_{\pm.007}$ \\
KID $\downarrow$       & $.263$ & $\mathbf{.096}_{\pm.013}$ \\
DINO sim $\uparrow$    & $.133$ & $\mathbf{.371}_{\pm.021}$ \\
CLIP DirSim $\uparrow$ & ---    & $.086_{\pm.006}$ \\
\midrule
\multicolumn{3}{l}{\textit{Retention}}\\
WikiText PPL $\downarrow$ & $13.71_{\pm.21}$ & $14.18_{\pm.22}$ \\
MMLU $\uparrow$           & $.492_{\pm.013}$ & $.479_{\pm.013}$ \\
POPE acc $\uparrow$       & $.877_{\pm.006}$ & $.871_{\pm.006}$ \\
MME $\uparrow$            & $1785_{\pm31}$   & $1713_{\pm31}$ \\
\bottomrule
\end{tabular}
\label{tab:sketch}
\end{table}

\section{Text-direction sweeps}
\label{app:textsweep}

\subsection{A domain and a language, full configuration}
\label{app:targetstable}

\begin{table}[htbp]
\caption{\textbf{A distant domain and a new language.} Both targets are learned strongly while every untrained capability stays at the base level. Both use the full evaluation configuration, so one base column serves both. Gen-CMMD is measured against the base model's own generations on $128$ neutral COCO prompts, not against real style images. Standard errors are given once per row; n/a marks metrics defined only for the other target.}
\label{tab:targets}
\centering\small
\setlength{\tabcolsep}{5pt}
\begin{tabular}{lccc}
\toprule
& Base (full) & Medicine LoRA & Welsh LoRA \\
&      & ($r{=}32$, $\lambda{=}0.25$) & ($r{=}32$, $\lambda{=}0.5$) \\
\midrule
\multicolumn{4}{l}{\emph{Acquisition}}\\
PubMed held-out PPL $\downarrow$        & 12.8 & \textbf{8.4} & n/a \\
PubMed PPL $\times$control $\downarrow$ & 1.23$\times$ & \textbf{0.79$\times$} & n/a \\
Welsh held-out PPL $\downarrow$         & 40.7 & n/a & \textbf{9.6} \\
$\lVert\Delta W\rVert/\lVert W\rVert$   & --- & 1.7\% & 1.9\% \\
\midrule
\multicolumn{4}{l}{\emph{Untrained capabilities}}\\
WikiText PPL $\downarrow$ \tiny($\pm.21$) & 13.71 & 13.97 & 12.16 \\
MMLU $\uparrow$ \tiny($\pm.013$)          & .492  & .495  & .490 \\
POPE acc $\uparrow$ \tiny($\pm.006$)      & .877  & .877  & .877 \\
MME $\uparrow$ \tiny($\pm31$)             & 1785  & 1766  & 1800 \\
\midrule
\multicolumn{4}{l}{\emph{Shift in the other modality}}\\
Gen-CMMD $\downarrow$ ($\times10^{-3}$) & 0 & 1.77 & \textbf{1.83} \\
\bottomrule
\end{tabular}
\end{table}

Tables~\ref{tab:textkl} and~\ref{tab:textrank} vary the two knobs on the poetry target:
the KL weight at fixed rank $16$, and the rank at fixed $\lambda{=}0.5$. Both come from
the main text, which quotes only their conclusions. Raising the rank from $4$ to $64$
steadily improves learning of the style. Poetry perplexity goes from $32.7$ to $23.8$.
MMLU, POPE and MME stay at the base level at every rank in the sweep. WikiText perplexity is not
degraded and is slightly below the base throughout. Specializing text therefore leaves
the other capabilities intact at every capacity, mirroring the image result of
Table~\ref{tab:rank}. Pushing the rank to $128$ confirms diminishing returns. The style
improves only marginally, poetry perplexity $23.8$ to $23.2$. A first small cost to
understanding appears: MME $1130$ to $1105$, POPE $0.875$ to $0.867$ against the base.
This is the second of the three convergent settings of Section~\ref{sec:boundary}. We
therefore report up to rank $64$, the point where the trade-off turns.

\begin{table}[htbp]
\caption{\textbf{Text (poetry): KL sweep at rank 16.} The style is learned best (bold) at low $\lambda$ and less well as $\lambda$ grows. \emph{Gen-CMMD} is the CMMD between the adapter's and the base model's generated images on $128$ fixed prompts: it stays near zero at every $\lambda$, so the text adapter does not measurably change image generation. Subsampled configuration; standard errors once per row.}
\label{tab:textkl}
\centering\small
\begin{tabular}{lcccccc}
\toprule
& Base & $\lambda{=}0$ & $\lambda{=}0.25$ & $\lambda{=}0.5$ & $\lambda{=}1$ & $\lambda{=}1.5$ \\
\midrule
\multicolumn{7}{l}{\emph{Acquisition} (target style)}\\
Poetry PPL $\downarrow$ & 32.7 & \textbf{24.5} & \textbf{24.5} & 24.9 & 25.9 & 26.7 \\
\midrule
\multicolumn{7}{l}{\emph{Untrained capabilities}}\\
WikiText PPL $\downarrow$ \tiny($\pm.21$) & 13.71 & 12.61 & \underline{12.50} & 12.51 & 12.75 & 12.93 \\
MMLU $\uparrow$ \tiny($\pm.021$)          & .481 & .474 & .481 & .479 & .475 & .483 \\
POPE acc $\uparrow$ \tiny($\pm.010$)      & .875 & .868 & .870 & .871 & .878 & .873 \\
MME $\uparrow$ \tiny($\pm42$)             & 1130 & 1101 & 1116 & 1128 & 1127 & 1128 \\
\midrule
\multicolumn{7}{l}{\emph{Shift in the other modality}}\\
Gen-CMMD $\downarrow$ ($\times10^{-3}$) & 0 & 1.73 & 1.71 & 1.73 & 1.77 & 1.65 \\
\bottomrule
\end{tabular}
\end{table}

\begin{table}[htbp]
\centering\small
\caption{Text (poetry) rank sweep at $\lambda{=}0.5$. The style is learned (bold = best) while knowledge and understanding stay flat; WikiText perplexity is slightly lower than the base model's at every rank (best underlined). Subsampled configuration, single seed.}
\setlength{\tabcolsep}{7pt}
\begin{tabular}{lcccccc}
\toprule
& Base & $r{=}4$ & $r{=}8$ & $r{=}16$ & $r{=}32$ & $r{=}64$ \\
\midrule
\multicolumn{7}{l}{\textit{Acquisition (target style)}}\\
Poetry PPL $\downarrow$ & $32.7$ & $25.6$ & $25.3$ & $24.9$ & $24.4$ & $\mathbf{23.8}$ \\
\midrule
\multicolumn{7}{l}{\textit{Untrained capabilities}}\\
WikiText PPL $\downarrow$ \tiny($\pm.21$) & $13.71$ & $12.52$ & $\underline{12.51}$ & $\underline{12.51}$ & $12.53$ & $12.59$ \\
MMLU $\uparrow$ \tiny($\pm.021$)          & $.481$ & $.481$ & $.477$ & $.479$ & $.477$ & $.488$ \\
POPE acc $\uparrow$ \tiny($\pm.010$)      & $.875$ & $.871$ & $.871$ & $.871$ & $.873$ & $.877$ \\
MME $\uparrow$ \tiny($\pm42$)             & $1130$ & $1129$ & $1111$ & $1128$ & $1130$ & $1122$ \\
\bottomrule
\end{tabular}
\label{tab:textrank}
\end{table}

\section{Size of the update and where it lands}
\label{app:dw}

\begin{figure}[htbp]
\centering
\begin{subfigure}{0.48\textwidth}
\includegraphics[width=\linewidth]{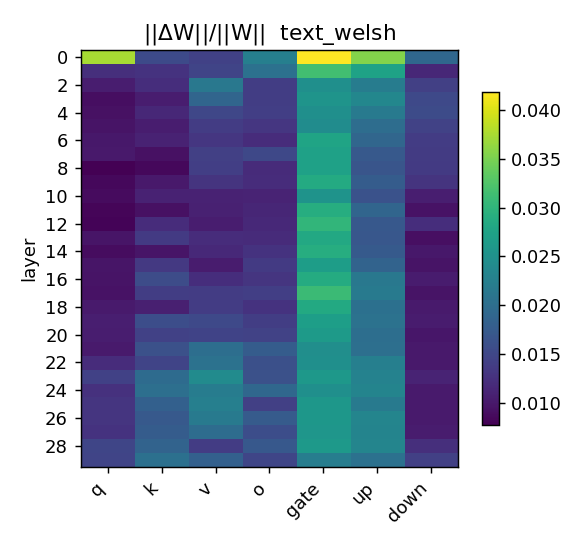}
\caption*{\footnotesize Text (Welsh, $r{=}32$)}
\end{subfigure}
\hfill
\begin{subfigure}{0.48\textwidth}
\includegraphics[width=\linewidth]
{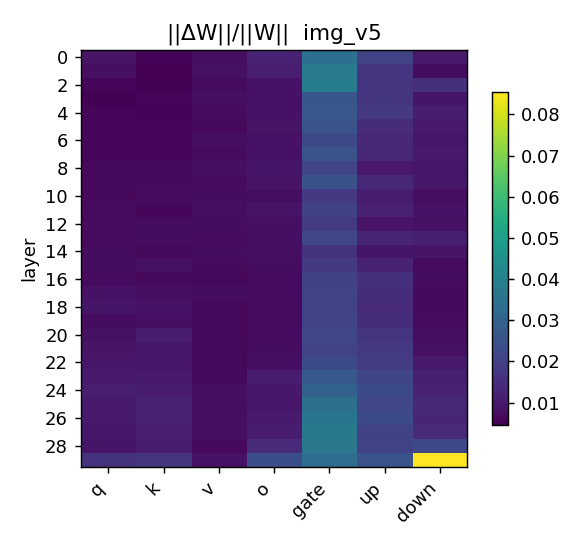}
\caption*{\footnotesize Image (anime, Final)}
\end{subfigure}
\caption{Size of the weight change per module, by layer (rows) and projection (columns); Table~\ref{tab:dw-proj} gives the per-projection means. \textbf{The two panels use different color scales}: the image panel reaches $0.085$, the text panel $0.042$. Apart from \texttt{gate} receiving the most change in both, the pattern over depth differs: the image adapter puts one very large change on \texttt{down} in the final layer, the text adapter an outlying first layer. Backbone modules only.}
\label{fig:dw-heatmap}
\end{figure}

Table~\ref{tab:dw-rank} gives the aggregate weight-change sizes quoted in
\S\ref{sec:dw}. The base norm $\lVert W\rVert_F$ is read from the frozen weights.
$\lVert\Delta W\rVert_F=\lVert(\alpha/r)BA\rVert_F$ is computed from the LoRA factors
through the $r\times r$ trace identity, so the full $\Delta W$ is never formed.
Table~\ref{tab:dw-proj} gives the same quantity per projection and
Figure~\ref{fig:dw-heatmap} per module. This makes the claim about \emph{where} the
change lands concrete instead of aggregate.

\begin{table}[htbp]\centering\small
\caption{Size of the weight change, $\lVert\Delta W\rVert_F/\lVert W\rVert_F$ (\%, summed over backbone projections), against LoRA rank (top) and against the KL weight (bottom). It grows smoothly with rank in both directions and generally decreases at larger KL weights, more sharply for text.}
\label{tab:dw-rank}
\begin{tabular}{lcccccc}
\toprule
\multicolumn{7}{l}{\emph{versus LoRA rank}}\\
Rank $r$         & 2    & 4    & 8    & 16   & 32   & 64 \\
\midrule
Image (gen.)     & 0.58 & 0.75 & 0.97 & 1.25 & 1.57 & 1.91 \\
Text (poetry)    & ---  & 0.67 & 0.84 & 1.05 & 1.32 & 1.66 \\
\midrule
\multicolumn{7}{l}{\emph{versus the KL weight $\lambda$}}\\
$\lambda$        & 0    & 0.25 & 0.5  & 1    & 1.5  & \\
\midrule
Image ($r{=}32$) & 1.55 & 1.58 & 1.57 & 1.45 & 1.33 & \\
Text ($r{=}16$)  & 1.15 & 1.15 & 1.05 & 0.89 & 0.80 & \\
\bottomrule
\end{tabular}
\end{table}

\begin{table}[htbp]\centering\small
\caption{Where the weight change lands: mean and maximum of $\lVert\Delta W\rVert_F/\lVert W\rVert_F$ (\%) over the $30$ backbone layers, per projection. \texttt{gate} receives the largest change in both directions, the only part of the pattern shared by the two: on average the image adapter puts $2.3$ times more change into the MLP than into attention, the text adapter only $1.4$, and for text \texttt{down} ranks below three of the four attention projections. SigLIP modules are excluded because they are never updated.}
\label{tab:dw-proj}
\begin{tabular}{lcccc}
\toprule
& \multicolumn{2}{c}{Text (Welsh, $r{=}32$)} & \multicolumn{2}{c}{Image (anime, $r{=}48$)} \\
\cmidrule(lr){2-3}\cmidrule(lr){4-5}
Projection & mean & max & mean & max \\
\midrule
\texttt{q}    & 1.15 & 3.73 & 0.81 & 1.65 \\
\texttt{k}    & 1.38 & 2.04 & 0.82 & 1.66 \\
\texttt{v}    & 1.58 & 2.42 & 0.71 & 0.85 \\
\texttt{o}    & 1.46 & 2.24 & 0.93 & 2.41 \\
\midrule
\texttt{gate} & \textbf{2.73} & 4.18 & \textbf{2.60} & 3.85 \\
\texttt{up}   & 2.10 & 3.55 & 1.65 & 2.54 \\
\texttt{down} & 1.18 & 1.91 & 1.27 & \textbf{8.55} \\
\bottomrule
\end{tabular}
\end{table}

\subsection{Regularizer ablation}
\label{app:reg}
We compare the KL-to-base regularizer at fixed rank 32 for image generation against
three alternatives: L2-SP \citep{xuhong2018l2sp} (an L2 penalty on the LoRA update, in
weight space), logit-MSE (a squared-error match to the base model's logits, in output
space), and the reconstruction-loss objective of Anole \citep{chern2024anole}, the
standard recipe for fine-tuning native generators (\texttt{CE}$+0.5\,$recon, no
reference model). All share the same training budget and data and are evaluated
identically on $80$ held-out prompts under the canonical, trigger-free protocol used in
the rest of the paper; we also report each adapter's weight-change size
$\lVert\Delta W\rVert_F/\lVert W\rVert_F$. The KL penalty learns the style best of all
methods. At essentially the same weight-change size as the unregularized baseline
($1.57$ vs.\ $1.55\%$) it reaches a markedly lower CMMD ($.108$ vs.\ $.155$), it
redirects the change rather than merely shrinking it. Logit-MSE over-constrains the
output distribution. It keeps image quality high but barely transfers the style without
a prompt trigger (DirSim $.029$, the worst of all methods), which the prompt-triggered
evaluation had hidden. Weight-space L2-SP simply shrinks the change, to $0.60\%$.

The untrained capabilities tell a different and more important story: they do not
distinguish between the objectives at all. Across all five, MME spans $31$ points,
exactly one standard error. MMLU, POPE and WikiText perplexity stay within noise of the
base model. Three of the five objectives produce the same weight-change size ($1.55$,
$1.56$, $1.57\%$). The objective is the only thing that differs between them, and it
makes no measurable difference to what is preserved. We therefore choose the KL penalty
for how well it learns. At this scale the locality reported throughout this paper is not
produced by it.

Two rows support that reading. L2-SP moves the weights least of all, $0.60\%$. It is the
only condition whose general-text perplexity shifts at all ($13.71\!\to\!14.05$, about
$1.6$ standard errors, so suggestive rather than established). Logit-MSE does the
opposite. It changes $1.14\%$ of the weight energy while transferring almost no style,
and costs nothing anywhere. Together they suggest that the direction of a weight change,
not its size, is what could threaten the other capabilities. These are single-seed runs.

\begin{table}[htbp]\centering\footnotesize
\caption{Regularizer comparison at rank $32$ (image direction, single seed; bold = best). The five objectives differ by a factor of two in how well they learn the style, from CMMD $.108$ to $.216$, and not at all in what they preserve: MME spans $31$ points across all five, one standard error. Only None, KL and Anole produce the same weight change and can be ranked against each other. $\lVert\Delta W\rVert$ is $\lVert\Delta W\rVert_F/\lVert W\rVert_F$ in \%. Standard errors: CMMD $\pm.010$, CLIP-IQA $\pm.011$, DirSim $\pm.007$; WikiText $\pm.21$, MMLU $\pm.013$, POPE $\pm.006$, MME $\pm31$.}
\label{tab:reg-ablation}
\setlength{\tabcolsep}{3pt}
\begin{tabular}{lcccccccc}
\toprule
& \multicolumn{4}{c}{\emph{Acquisition} (vs.\ real style, $N{=}80$)}
& \multicolumn{4}{c}{\emph{Untrained capabilities}} \\
\cmidrule(lr){2-5}\cmidrule(lr){6-9}
Regularizer & $\lVert\Delta W\rVert$ & CMMD $\downarrow$ & CLIP-IQA $\uparrow$ & DirSim $\uparrow$
            & WikiText $\downarrow$ & MMLU $\uparrow$ & POPE $\uparrow$ & MME $\uparrow$ \\
\midrule
\emph{base (no adapter)} & --- & --- & --- & --- & 13.71 & .492 & .877 & 1785 \\
\midrule
None ($\lambda{=}0$) & 1.55 & .155 & .666 & .157 & 13.67 & .495 & .879 & 1783 \\
L2-SP                & 0.60 & .139 & .683 & \textbf{.164} & 14.05 & .485 & .878 & 1781 \\
Logit-MSE            & 1.14 & .216 & \textbf{.725} & .029 & 13.70 & .496 & .876 & 1798 \\
KL-to-base (ours)    & 1.57 & \textbf{.108} & .691 & .153 & 13.78 & .497 & .879 & 1796 \\
Anole (recon)        & 1.56 & .148 & .666 & .162 & 13.74 & .498 & .876 & 1767 \\
\bottomrule
\end{tabular}
\end{table}

\section{Qualitative examples}
\label{app:qual}

\subsection{Text-direction continuations}
\label{app:textqual}\label{app:medicine}\label{app:welsh}
Base-versus-adapter continuations for the three text targets. In each pair the base is
the same model with the adapter switched off, so the columns differ only by the LoRA.
Decoding is greedy with a repetition penalty of $1.3$ (Appendix~\ref{app:impl}). Poetry
openings are continued from a fixed verse couplet given as hidden context to \emph{both}
models. This keeps the unified model in a poem rather than its assistant persona, and
only the text after the opening is shown.

\begin{table}[htbp]
\centering\footnotesize
\caption{Text continuations for the three targets. Poetry keeps lineated verse where the base model drifts into prose or scriptural citation; medicine gives clinical content in PubMed register while both models answer the general-knowledge control correctly, so the domain adaptation costs no factual recall; Welsh is continued from a Welsh opening and \emph{not} from an English one, so the language is added without overwriting the existing one. The leftover ``Hello!\dots'' in the base column appears because Janus-Pro is partly instruction-tuned.}
\setlength{\tabcolsep}{5pt}\renewcommand{\arraystretch}{1.0}
\begin{tabular}{@{}p{0.46\linewidth}p{0.46\linewidth}@{}}
\toprule
\textbf{Base (adapter off)} & \textbf{Adapter} \\
\midrule
\multicolumn{2}{@{}l@{}}{\emph{Poetry LoRA} \quad \textit{incipit:} ``Upon the silent hill there\,\dots''}\\
lies a light of gold; ,  O! how it gleams in twilight's shade: it is not day nor
night but something more divine.
&
stands a tree;\newline
The leaves are green as emeralds in their sleep:\newline
And on its branches hang an empty cage,\newline
A prison for some bird of sorrow's keep!\\
\addlinespace[3pt]
\multicolumn{2}{@{}l@{}}{\emph{Poetry LoRA} \quad \textit{incipit:} ``O gentle heart, what\,\dots''}\\
words of love can say? --- O! for a thousand tongues to tell thee how I feel this
day; but none are fit nor true \dots\ (1 Samuel 25).
&
is thy secret? What art thou made of but a little child's delight in play; \dots\
O say not ``I am nothing,'' for I see how much you have to give\,\dots\\
\midrule
\multicolumn{2}{@{}l@{}}{\emph{Medicine LoRA} \quad ``What is the mechanism of action of metformin?''}\\
Metformin works by inhibiting glucose production in cells, particularly those
that are most active at high concentrations. Hello! I'm here to help you with
your question [\dots]
&
Metformin works by inhibiting glucose uptake in muscle cells and reducing hepatic
gluconeogenesis. [\dots]\\
\addlinespace[3pt]
\multicolumn{2}{@{}l@{}}{\emph{Medicine LoRA} \quad \textit{control:} ``Who wrote the play Romeo and Juliet?''}\\
William Shakespeare. [\dots] the playwright of this famous tragedy [\dots]
($1564$--$1623$).
&
William Shakespeare. [\dots] a famous English playwright who lived between
approximately $1564$--$1616$ [\dots] including Romeo \& Juliet!\\
\midrule
\multicolumn{2}{@{}l@{}}{\emph{Welsh LoRA} \quad \textit{Welsh seed:} ``Mae Cymru yn wlad\,\dots''}\\
fod â'r eich, a dda i gael \dots\ Hello Mae! I am happy to help you with your
Welsh language question.
&
fwyaf o'r gogledd-ddinasol a ddau 10,578.234 iawn (a chyfansoddwyd gan yr Aelodau
Canolfan Llyfrau Genedlaethol). Mae hynny wedi ei bod arall am\,\dots\\
\addlinespace[3pt]
\multicolumn{2}{@{}l@{}}{\emph{Welsh LoRA} \quad \textit{English seed:} ``The history of Wales\,\dots''}\\
is a long and complex one, with many different cultures influencing its
development over the centuries. Hello!
&
is a long and complex one, with many different cultures influencing the country
over time. I will be writing about some key events in Welsh History that have
shaped our nation today\,\dots\\
\bottomrule
\end{tabular}
\label{tab:text-examples}
\end{table}

\begin{figure}[htbp]\centering
\setlength{\tabcolsep}{1.5pt}\renewcommand{\arraystretch}{1.0}\small

\textbf{(a) English prompts: the adapter changes nothing}\\[2pt]
\begin{tabular}{@{}cccc@{}}
\rotatebox{90}{\scriptsize\ \ Base} &
\includegraphics[width=0.24\linewidth]{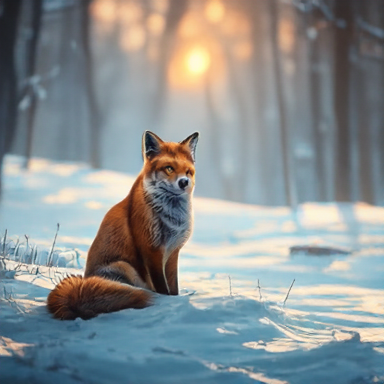} &
\includegraphics[width=0.24\linewidth]{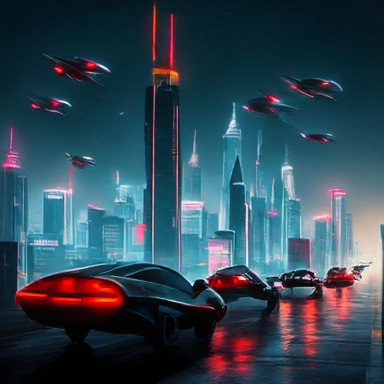} &
\includegraphics[width=0.24\linewidth]{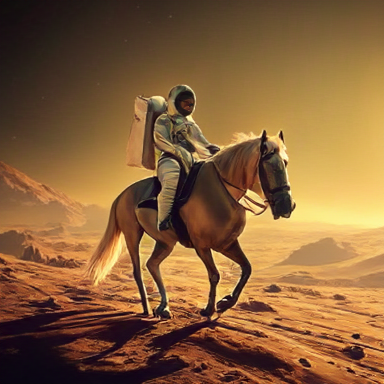} \\
\rotatebox{90}{\scriptsize Welsh LoRA} &
\includegraphics[width=0.24\linewidth]{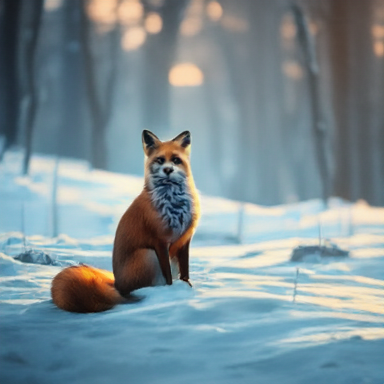} &
\includegraphics[width=0.24\linewidth]{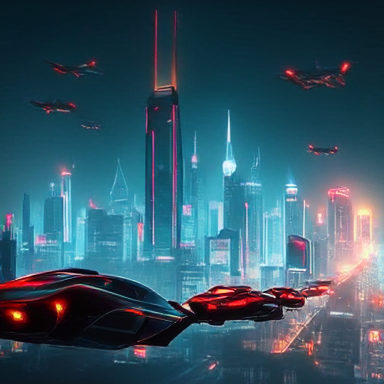} &
\includegraphics[width=0.24\linewidth]{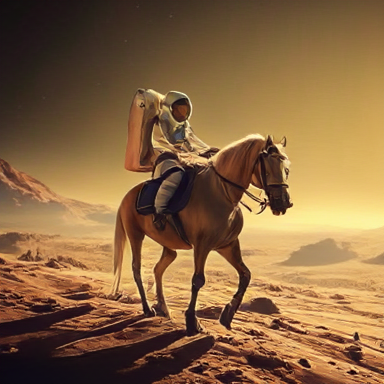} \\
&
\scriptsize\textit{red fox in snow} &
\scriptsize\textit{city skyline} &
\scriptsize\textit{astronaut on Mars} \\
\end{tabular}

\vspace{6pt}
\textbf{(b) Welsh prompts: neither base nor adapter renders the content}\\[2pt]
\begin{tabular}{@{}ccc@{}}
\scriptsize English $\to$ Base & \scriptsize Welsh $\to$ Base & \scriptsize Welsh $\to$ Welsh LoRA \\
\includegraphics[width=0.24\linewidth]{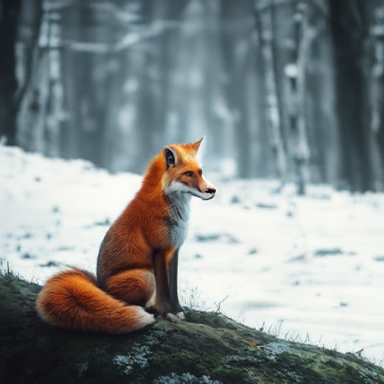} &
\includegraphics[width=0.24\linewidth]{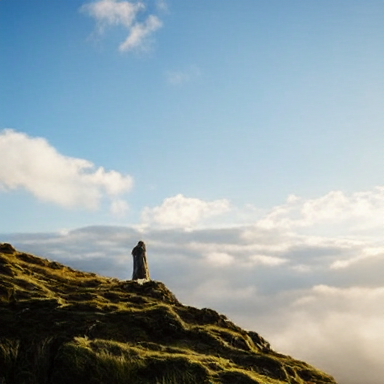} &
\includegraphics[width=0.24\linewidth]{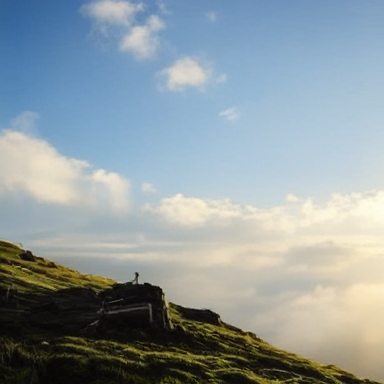} \\
\multicolumn{3}{@{}l}{\scriptsize\textit{a red fox / llwynog coch yn eistedd mewn coedwig eira}} \\
\addlinespace[2pt]
\includegraphics[width=0.24\linewidth]{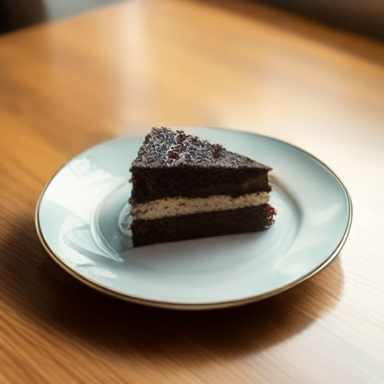} &
\includegraphics[width=0.24\linewidth]{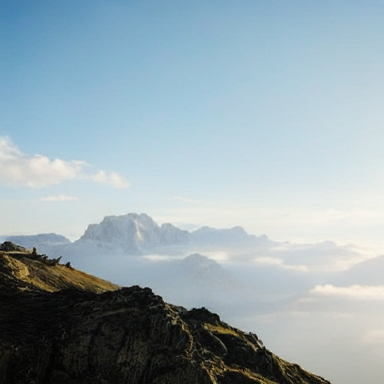} &
\includegraphics[width=0.24\linewidth]{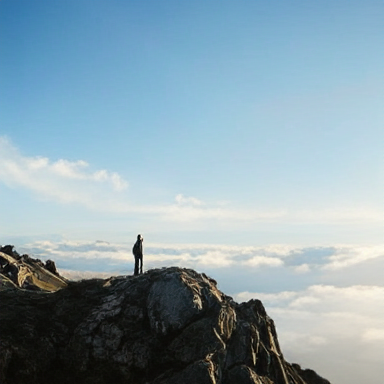} \\
\multicolumn{3}{@{}l}{\scriptsize\textit{a slice of chocolate cake / tafell o gacen siocled ar bl\^at}} \\
\end{tabular}

\caption{The two sides of locality, same seeds throughout.
\textbf{(a)} With English prompts the Welsh adapter draws what the base draws:
Gen-CMMD $1.8\times10^{-3}$ (Table~\ref{tab:targets}), against $18.7\times10^{-3}$ for an
adapter that does change the output.
\textbf{(b)} With Welsh prompts neither model renders the content named, and the adapter
fails in the same way as the base ($+0.48\pm0.24$ CLIPScore, Table~\ref{tab:xmodal}). The
adapter neither damages image generation nor reaches it.}
\label{fig:welsh-visual}\label{fig:t2i_welsh}\label{fig:xmodal-welsh}
\end{figure}

\subsection{Free-form English and MAUVE}
\label{app:mauve}
Table~\ref{tab:mauve} gives the full MAUVE panel quoted in \S\ref{sec:text}. The base
model has a low absolute MAUVE against WikiText ($0.38$). This is not a sign of broken
output. Its continuations are coherent encyclopedic prose, as in
``\emph{Mortier accepted the bait of a rumored Russian retreat.}
\emph{In the early morning of 11 November he and Gazan departed from D\"urenstein to seize the abandoned fort\dots}''
The low value reflects the gap between the text distribution of a multimodal assistant
model and encyclopedic text. MAUVE depends on our fixed protocol ($N{\le}2000$,
GPT-2-large features). It is meaningful within the table, not as an absolute number to
compare across papers.

\begin{table}[htbp]
\caption{\textbf{Free-form generation (MAUVE)} ($[0,1]$, higher = closer; $N\!\le\!2000$, GPT-2-large features). The image adapters are indistinguishable from the base model in English; no adapter is farther from human text than the base model; the text adapters move much closer to their targets. MAUVE has no closed-form standard error, so we read only differences larger than the $0.1$ steps on which these entries differ.}
\label{tab:mauve}
\centering\small
\begin{tabular}{lccc}
\toprule
Adapter & English & English & Target acq. \\
        & vs.\ human & vs.\ base & base\,$\to$\,adapter \\
\midrule
Base (reference)        & 0.38 & --- & --- \\
Anime, image (Final)    & 0.52 & 0.95 & --- \\
Sketch, image           & 0.46 & 0.92 & --- \\
Poetry                  & 0.70 & 0.83 & 0.45\,$\to$\,0.80 \\
Medicine                & 0.60 & 0.21 & 0.31\,$\to$\,0.91 \\
Welsh                   & 0.63 & 0.79 & 0.14\,$\to$\,0.53 \\
\bottomrule
\end{tabular}
\end{table}

\section{Cross-modal probes: protocol}
\label{app:xmodal}
The probes in Table~\ref{tab:xmodal} are built as follows. For (A) we generate the same
content from $30$ parallel English/Welsh prompt pairs. Examples are
\emph{a yellow school bus on a city street} / \emph{bws ysgol melyn ar stryd y ddinas},
or \emph{a plate of fish and chips} / \emph{pl\^at o bysgod a sglodion}, with two
further pairs shown under Figure~\ref{fig:xmodal-welsh}. Every image is scored against
the same \emph{English} reference with CLIPScore using a strong English CLIP. This means
the Welsh side is scored on meaning rather than by a text encoder that cannot read
Welsh. For (B) the image and the answer format are fixed and only the language of the
question changes. Answers are scored by the yes/no logit, so drift in the output
language cannot corrupt them. Because the base model barely knows Welsh, an improvement
would be transfer of the new language across modalities, not a retention effect.

\begin{table}[htbp]
\caption{\textbf{Does the learned language reach the visual pathway?} The English controls match the base model, so the probes work. On Welsh the adapter improves image generation from Welsh prompts only marginally ($+0.48\pm0.24$ over $30$ paired prompts) and visual question answering not at all; both VQA scores lie inside the Wilson interval of chance. What was learned from text alone does not reach the visual pathway.}
\label{tab:xmodal}
\centering\small
\begin{tabular}{lcccc}
\toprule
& \multicolumn{2}{c}{Base} & \multicolumn{2}{c}{Welsh LoRA} \\
\cmidrule(lr){2-3}\cmidrule(lr){4-5}
& English & Welsh & English & Welsh \\
\midrule
(A) text$\to$image, CLIPScore $\uparrow$ ($n{=}30$) & 26.34 & 10.66 & 26.51 & 11.14 \\
(B) VQA accuracy $\uparrow$ ($n{=}28$)              & 100\% & 53.6\% & 100\% & 46.4\% \\
\quad\tiny 95\% Wilson interval & \tiny[88,100] & \tiny[36,70] & \tiny[88,100] & \tiny[30,64] \\
\bottomrule
\end{tabular}
\end{table}

One validity check on (A) is whether Welsh text reaches the generator. If it never did,
the Welsh columns would measure the model's unconditional output and the comparison
would be empty. With the seed fixed and only the prompt varied, three Welsh prompts with
unrelated content produce three different images. Each of those images also differs from
the image generated from an empty prompt, so the text does condition generation. The
empty prompt yields a generic landscape, and every Welsh prompt stays close to it. It
does not move toward the fox, the cake or the bus it names. The probe therefore measures
a failure to convey content, with conditioning present but ineffective.

Probe~(B) fails for a reason visible at the item level. On the Welsh items the base model
answers ``yes'' to $11$ of the $14$ questions whose correct answer is ``no'', and gets
$12$ of the $14$ ``yes'' items right. It is not reading the Welsh question at all, and its
yes-bias turns a balanced set into chance-level accuracy. A probe the model cannot pass in
principle measures the joint failure of Welsh understanding and Welsh instruction
following, not the absence of cross-modal transfer.

\end{document}